\documentclass[11pt]{article}

\usepackage[final]{acl}

\usepackage{times}
\usepackage{latexsym}

\usepackage[T1]{fontenc}
\usepackage[utf8]{inputenc}

\usepackage{microtype}

\usepackage{inconsolata}

\usepackage{graphicx}

\usepackage{amssymb}

\usepackage{booktabs}
\usepackage{amsthm}
\newtheorem{definition}{Definition}
\usepackage{graphicx} % 绘制图像
\usepackage{enumitem} % 缩进
\usepackage{bbding}
\usepackage{multirow}
\usepackage{makecell}
\usepackage{subfigure}
\usepackage{array} % for case study

\usepackage{algorithm}
\usepackage{algorithmic}

\usepackage[most]{tcolorbox}
\usepackage{tcolorbox}
\usepackage[table,xcdraw]{xcolor}  % 加载xcolor宏包
\definecolor{mybackcolor}{RGB}{240, 246, 232}
\definecolor{myframecolor}{RGB}{102,166,30}
\definecolor{mybackcolor1}{RGB}{242,242,242}
\definecolor{myframecolor1}{RGB}{219,219,219}
\usepackage{lipsum} % For placeholder text
\tcbset{
    mytakeaway/.style={
        enhanced,
        breakable,  % Allow box to break across pages or columns
        colback=mybackcolor,
        colframe=myframecolor,
        left=2mm,
        right=2mm,
        boxrule=0mm,
        arc=0mm,
        width=\columnwidth,
        before skip=10pt,
        after skip=10pt,
        frame hidden, % Hide all borders
        borderline west={1mm}{0mm}{myframecolor}, % Add left border
    }
}
\tcbuselibrary{listings} % Load the listings library for labeling
\newtheorem{prompt}{Prompt}
\tcbset{
    prompt/.style={
        enhanced,
        breakable,  % Allow box to break across pages or columns
        colback=mybackcolor1,
        colframe=myframecolor1,
        left=2mm,
        right=2mm,
        boxrule=0mm,
        arc=0mm,
        before skip=10pt,
        after skip=10pt,
        frame hidden, % Hide all borders
        borderline west={0mm}{0mm}{myframecolor1}, % Add left border
        title=#1, % Title with prompt number
        coltitle=black, % Set title color
        fonttitle=\bfseries, % Set title font
        colbacktitle=myframecolor1, % Set title background color
        width=\columnwidth, % Span across single column width
        left=0pt, % Reset any side margins to avoid shifting
        right=0pt, % Reset right margin
    }
}

\title{Joint Knowledge Base Completion and Question Answering by Combining Large Language Models and Small Language Models}

\author{
 \textbf{Yinan Liu$^1$}\thanks{\ \ Equal contribution.}\thanks{\ \ Corresponding authors.},
 \textbf{Dongying Lin$^1$}\footnotemark[1],
 \textbf{Sigang Luo$^1$},
 \textbf{Xiaochun Yang$^{1,2}$},
 \textbf{Bin Wang$^{1,2}$}
\\
$^1$School of Computer Science and Engineering, Northeastern University, Shenyang, China\\
$^2$National Frontiers Science Center for Industrial Intelligence and Systems optimization, \\Northeastern University, Shenyang, China\\
\small \texttt{liuyinan@cse.neu.edu.cn},
\small \texttt{2472128@stu.neu.edu.cn}, \\
\small \texttt{2301925@stu.neu.edu.cn},
\small \texttt{\{yangxc, binwang\}@mail.neu.edu.cn}
}

\begin{document}
\maketitle
\begin{abstract}
Knowledge Bases (KBs) play a key role in various applications. As two representative KB-related tasks, knowledge base completion (KBC) and knowledge base question answering (KBQA) are closely related and inherently complementary with each other. Thus, it will be beneficial to solve the task of joint KBC and KBQA to make them reinforce each other. However, existing studies usually rely on the small language model (SLM) to enhance them jointly, and the large language model (LLM)'s strong reasoning ability is ignored. In this paper, by combining the strengths of the LLM with the SLM, we propose a novel framework JCQL, which can make these two tasks enhance each other in an iterative manner. To make KBC enhance KBQA, we augment the LLM agent-based KBQA model's reasoning paths by incorporating an SLM-trained KBC model as an action of the agent, alleviating the LLM's hallucination and high computational costs issue in KBQA. To make KBQA enhance KBC, we incrementally fine-tune the KBC model by leveraging KBQA's reasoning paths as its supplementary training data, improving the ability of the SLM in KBC. Extensive experiments over two public benchmark data sets demonstrate that JCQL surpasses all baselines for both KBC and KBQA tasks.
\end{abstract}

\section{Introduction}
Knowledge Bases (KBs) storing enormous factual triples such as Wikidata \cite{vrandevcic2014wikidata}, DBpedia \cite{lehmann2015dbpedia}, and Freebase \cite{10.1145/1376616.1376746}, play a key role in various applications \cite{diaglink}. One of their main downstream applications is knowledge base question answering (KBQA), which is a significant research area that utilizes KBs to provide precise answers to natural language questions. However, these KBs are often greatly incomplete \cite{bordes2013}. Along this line, the task of knowledge base completion (KBC), which aims to infer the missing triples for a KB, can help to improve the performance of KBQA. Meanwhile, KBQA can help KBC since the new knowledge inferred from KBQA can be used to complete the KB. It can be seen that KBC and KBQA are closely related and inherently complementary with each other \cite{lihui2022binet}. Thus, it will be beneficial to solve KBC and KBQA jointly to make them reinforce each other.

 To our best knowledge, only two existing studies have attempted to integrate KBC and KBQA tasks by leveraging the small language models (SLMs) such as BERT \cite{DBLP:conf/naacl/DevlinCLT19} and T5 \cite{T5}. The first approach BiNet \cite{lihui2022binet} proposes a multi-task learning framework that shares an embedding space to facilitate latent feature exchange between these two tasks. However, this method requires manual annotation of reasoning paths (i.e., relation sequences of questions \cite{lihui2022binet}) in the embedding space to construct training data, which is time-consuming and labor-intensive. The second work, KGT5 \cite{saxena2022kgt5} formulates KBC and KBQA as seq2seq tasks. Specifically, KGT5 first pre-trains T5 for KBC and then fine-tunes it for KBQA, enabling knowledge integration between tasks through shared parameters. 
 While this method is simple and intuitive, it overlooks the potential enhancing effect of KBQA on KBC. Notably, it is worth mentioning that the previous two studies are both based on SLMs, failing to use the abilities of large language models (LLMs) such as GPT-4 \cite{achiam2023gpt} and LLaMA \cite{touvron2023llama}. These LLMs exhibit superior reasoning abilities that can retrieve relevant triples from the KB to generate the needed reasoning paths through exploration and reasoning \cite{sun2023thinkongraph}. A heuristic way to integrate KBC and KBQA tasks employing the LLM is establishing an iterative workflow, e.g., first executing an LLM-based KBC method to complete the KB, then running an LLM-based KBQA method, and subsequently leveraging newly acquired knowledge from KBQA's reasoning paths to populate the KB, which can be used as the input of the next round of KBC. However, this workflow suffers from two limitations: (1) the cost overhead from frequent LLM calls; (2) results provided by LLMs are not precise due to the LLM's hallucination issue.

The above analysis reveals fundamental limitations in using either SLMs or LLMs in isolation for solving KBC and KBQA jointly. Interestingly, we observe complementary strengths: LLMs can potentially address BiNet's manual annotation requirement through their reasoning abilities, while SLMs can help mitigate LLMs' high computational costs and hallucination issues. Therefore, an effective integration strategy to combine the strengths of LLMs and SLMs can better enable mutual enhancement between the two tasks \cite{DBLP:journals/pvldb/FanGZZCCLMDT24}.

\begin{figure*}[t!]
  \centering
  \includegraphics[width=0.96\linewidth]{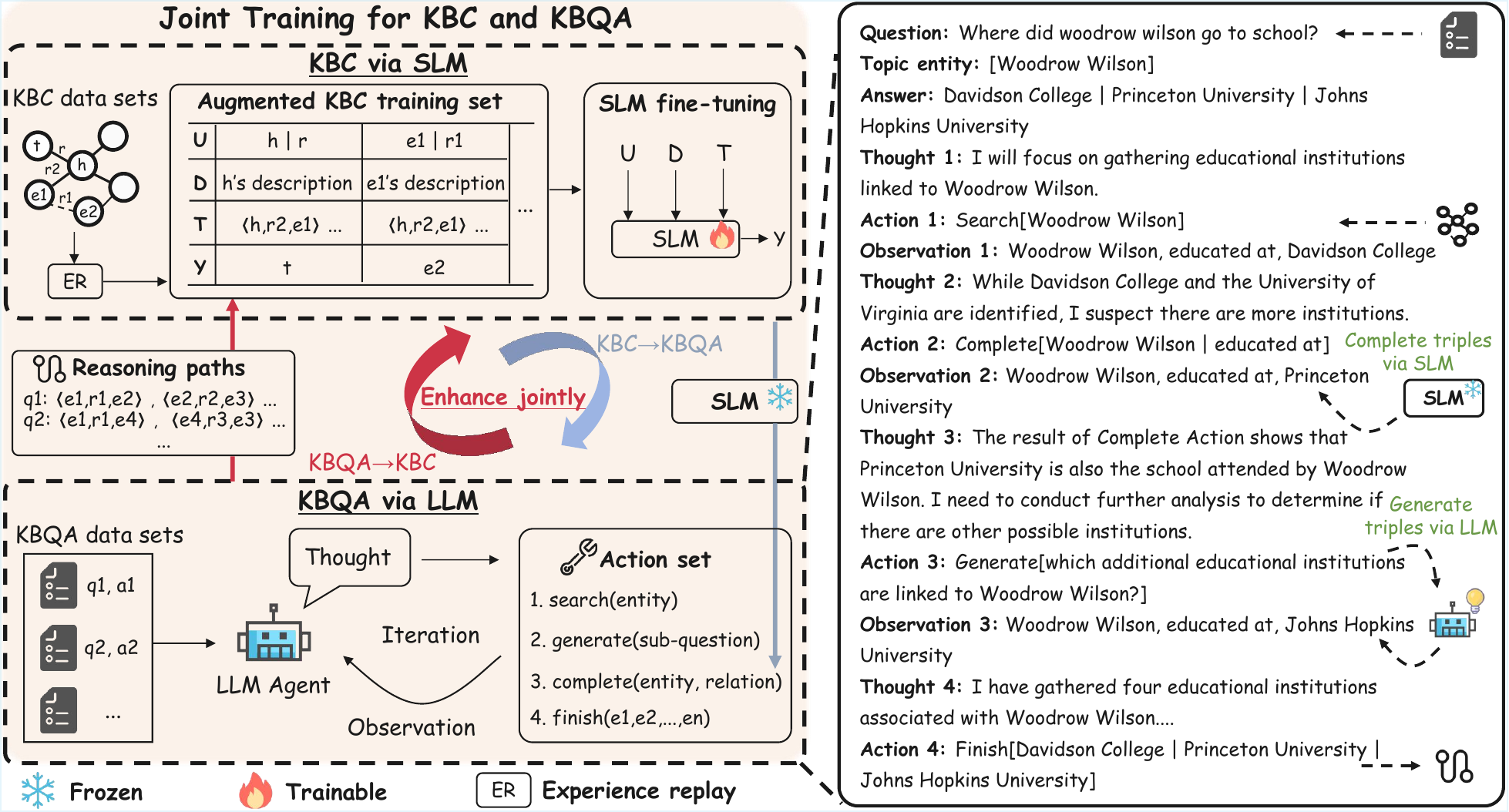}
    % \setlength{\abovecaptionskip}{-0.5mm}
    % \setlength{\belowcaptionskip}{-2mm}
  % \vspace{-4pt}
  \caption{Overview of our framework JCQL.}
  % \vspace{-4pt}
  \label{Framework overview}
\end{figure*}

% subgraph
% \begin{figure*}[htb!]
%     \centering
%     \subfigure[Fine-tune]{
%         \label{model overview:subfig:onefunction}
%         \includegraphics[width=0.56\textwidth]{image/img/Model_overview.pdf}}
%     \hspace{0.2em}
%     \subfigure[Inference]{
%         \label{model overview:subfig:twofunction}
%         \includegraphics[width=0.41\textwidth]{image/img/Inference.pdf}}
%     \caption{Model overview}\label{fig:init_mode}
%     \label{model overview}
% \vspace{-0.5em}
% \end{figure*}

To address the above issues, we propose a novel framework JCQL to \textbf{\underline{J}}ointly solve KB\textbf{\underline{C}} and KB\textbf{\underline{Q}}A by combining the \textbf{\underline{L}}LM and the SLM, which can integrate LLM's implicit parametric knowledge and KB’s explicit structured knowledge. For KBC, JCQL trains an SLM by retrieving relevant triples from the KB for each triple as contextual information. For KBQA, for each question, JCQL extracts the LLM's reasoning paths by using the corresponding answer as the supervision to prompt the LLM to act as an agent in step-by-step reasoning. In order to integrate KBC and KBQA into a unified framework to enhance each other, JCQL employs two key mechanisms: (1) the SLM-trained KBC model is integrated as an action within the LLM agent’s reasoning process, converting structured knowledge into parametric knowledge to alleviate the LLM’s hallucination and high computational costs issue in KBQA (a.k.a KBC $\rightarrow$ KBQA); (2) KBQA's reasoning paths are utilized as the supplementary training data to fine-tune the KBC model based on the incremental learning technique, converting parametric knowledge into structured knowledge to improve the ability of the SLM in KBC (a.k.a KBQA $\rightarrow$ KBC). Through the above iterative process, the KBC model augmented by KBQA's reasoning paths becomes stronger, and KBQA's reasoning paths incorporating the KBC model are more likely to be high quality. Ultimately, JCQL can predict entities via the fine-tuned SLM and generate answers to questions via the LLM agent incorporating the fine-tuned SLM as an action, respectively. 

Our main contributions can be summarized as follows: (1) to the best of our knowledge, this is the first framework combining the LLM and the SLM to enhance the task of joint KBC and KBQA; (2) to make KBC enhance KBQA, we augment the LLM agent by incorporating an SLM-trained KBC model as an action, which can alleviate the LLM's hallucination and high computational costs issue in KBQA; (3) to make KBQA enhance KBC, we incrementally fine-tune the KBC model by leveraging LLM-generated reasoning paths from KBQA as supplementary training data, which can improve the ability of the SLM in KBC; (4) a thorough experimental study over two public benchmark data sets under different settings shows our proposed framework significantly outperforms all baseline methods for both tasks.

\section{Preliminaries and Task Definition}
A KB can be represented as $G=\{E,R,F\}$, where $E$ is the set of entities, $R$ is the set of relations, and $F=\{\langle h,r,t\rangle|h,t \in E, r \in R\}$ is the set of triples. 
% A fact in a KB is denoted by $(h,r,t)$, where $h$ is the head entity, $t$ is the tail entity, and $r$ is the relation from the head entity to the tail entity.
% Moreover, there are some available text attributes in the KB such as the entity name, the relation name, and the entity description.
Given a KB, KBC aims to infer missing triples based on the existing information in the KB. Specifically, KBC consists of three categories of sub-tasks: (1) given a query $\langle h,r,?\rangle$, complete the corresponding tail entity $t$; (2) given a query $\langle ?,r,t\rangle$, complete the corresponding head entity $h$; (3) given a query $\langle h,?,t\rangle$, complete the relation $r$ between $h$ and $t$.
% Given a question, we refer to the entities mentioned in it as topic entities. 
KBQA aims to identify a set of entities $A_Q \subseteq E$ to answer the given questions $Q$ based on $G$. Following \cite{lihui2022binet}, we call the entities mentioned in questions as topic entities, denoted as $E_Q \subseteq E$. Ideally, each question $q \in Q$ can be mapped to a set of reasoning paths $P_q$ in the KB, where $p_q=\{\langle e_1,r_1,e_2\rangle, \langle e_2,r_2,e_3\rangle, \dots, \langle e_{d-1},r_{d-1},e_d\rangle\}$ is a unique reasoning path in $P_q$. 
% Ideally, each question $q \in Q$ can be mapped to a unique reasoning path $p_q=\{\langle e_1,r_1,e_2\rangle, \langle e_2,r_2,e_3\rangle, \dots, \langle e_{d-1},r_{d-1},e_d\rangle\} \subseteq P_q$ in the KB, where $P_q$ is the set of reasoning paths of $q$. 
% Notations used in this paper are summarized in Appendix \ref{appendix:notations and definitions}.
\begin{definition}[\textbf{Joint knowledge base completion and question answering}] 
Given a set of training triples for the KBC task and a set of training questions for the KBQA task, it is required to output answers to test triples for the KBC task and answers to test questions for the KBQA task.
\end{definition}

\section{The Framework: JCQL}
The overall framework JCQL is shown in Figure \ref{Framework overview}.
% and its each part will be elaborated as follows.

% \vspace{-1mm}
\subsection{KBC via SLM} \label{sec:3.1}
% \vspace{-1mm}
Previous studies \cite{saxena2022kgt5, yao2019kgbertbertknowledgegraph} demonstrate that SLMs can complete missing triples by learning semantic information from KBs. Inspired by this, we train an encoder-decoder Transformer model (with a similar architecture as T5) using the cross-entropy as the loss function to solve the KBC task. Specifically, the generation process of this SLM can be formalized as follows:
\begin{equation}
    \mathcal{P}(Y|[\mathcal{U}; \mathcal{D}; \mathcal{T}])=\prod_{i=1}^{l}{\mathcal{P}(y_i|y_{<i},[\mathcal{U}; \mathcal{D}; \mathcal{T}])},
\end{equation}
where $l$ is the length of sequence, $[\mathcal{U}; \mathcal{D}; \mathcal{T}]$ is the input sequence to the encoder of the SLM for each training triple $\langle h,r,t\rangle$, "$;$" is the concatenation operation between texts, $\mathcal{U}$ denotes the concatenation of $h$ and $r$, and $Y$ is the surface form of $t$. 
% Some studies \cite{yao2024exploringlargelanguagemodels, saxena2020improving} convert the query into a one-hop question, which yields good performance. 
% Inspired by \cite{yao2024exploringlargelanguagemodels, saxena2020improving}, we randomly sample some triples for $r$, and prompt the LLM to generate a template for $r$ with the sampled triples. Using these templates, $\langle h,r,?\rangle$ will be transformed into a one-hop question as $\mathcal{U}$. 
We obtain $h$'s descriptive text as \textbf{$\mathcal{D}$} from the KB, promoting the SLM to understand the entity better and alleviate the entity ambiguity issue. \textbf{$\mathcal{T}$} is the context consisting of related triples to $\mathcal{U}$.
% , which can provide more context information for the SLM to improve the performance furthermore. 
% Inspired by \cite{liu2024finetuninggenerativelargelanguage}, to mitigate redundant information and overly complex inputs, we propose a simple and efficient sampling strategy to capture explicit knowledge from the KB by incorporating the entity neighborhood information as the context compared with traditional KBC methods. 
% Specifically, we sample some triples consisting of $h$ with its one-hop neighborhoods as \textbf{$\mathcal{T}$} uniformly, at random, and without replacement. 
% \textcolor{red}{Specifically, we sample some triples consisting of $h$ with its one-hop neighborhoods as \textbf{$\mathcal{T}$}, which are ranked according to the co-occurrence score between $r$ and the neighbor relation $\hat{r}$:
% \begin{equation}
%     \mathcal{S}(r, \hat{r}) = \sum_{e \in \mathcal{E}} \mathbb{I}(r \in R_e \land \hat{r} \in R_e)
% \end{equation}
% where $R_e$ denotes the set of relations related to entity $e$, and $\mathbb{I}(\cdot)$ is the indicator function.} 
Specifically, we rank the one-hop neighborhoods of $h$ based on a co-occurrence score $\mathcal{S}(r, \hat{r})$ between $r$ and neighboring relations $\hat{r}$ and subsequently select the triples with high scores to construct the context set $\mathcal{T}$. Formally, $\mathcal{S}(r, \hat{r})$ is defined as follows:
\begin{equation}
    \mathcal{S}(r, \hat{r}) = \sum_{e \in \mathcal{E}} \mathbb{I}(r \in R_e \land \hat{r} \in R_e),
\end{equation}
where $R_e$ denotes the set of all relations incident to entity $e \in \mathcal{E}$. $\mathbb{I}(\cdot)$ denotes the indicator function and $\wedge$ denotes the logical conjunction. This score represents the frequency of $r$ and $\hat{r}$ sharing a common entity in the KB.
\subsection{KBQA via LLM} \label{sec:3.2}
% \vspace{-1mm}
Inspired by \cite{sun2023thinkongraph}, we utilize an LLM as the agent for KBQA. The LLM operates in an iterative cycle of three alternating states, i.e., \texttt{thought}, \texttt{action}, and \texttt{observation} to drive autonomous reasoning. During the reasoning process, the LLM continuously explores the KB and generates the reasoning process, dynamically integrating structured knowledge from the KB. To enable the LLM to access the KB and alleviate the KB's incomplete issue, we define the agent's action space consisting of the following three categories of actions: (1) \textbf{\texttt{search(entity)}}. This action can return the most relevant entities from the neighbors of the target entity based on some relevant relations from all triples associated with the target entity in the KB selected by the LLM (detailed in Appendix). (2) \textbf{\texttt{generate(sub-question)}}. 
The LLM is prompted to analyze the last \texttt{thought} to generate a sub-question. 
Based on the sub-question, this action retrieves related triples via BM25 algorithm \cite{robertson2009} from previous observations (i.e., action execution results), and then prompts the LLM to generate new triples beyond the KB via the LLM's parametric knowledge. Yet, the LLM may frequently return unreliable triples (e.g., inverted head-tail entity pairs and synthesized relationships unsupported by the KB). To solve this issue, JCQL prompts the LLM to correct these triples based on the description and schema of relationships (detailed in Appendix).
% \ref{appendix:prompt:supplementary}
 (3) \textbf{\texttt{finish($e_1,e_2,\dots,e_n$)}}. This action returns an entity list as final answers, marking the completion of the reasoning process. 

% \begin{itemize}[left=0pt]
%     \item \textbf{\texttt{search(entity)}}. This action can return the most relevant entities from the neighbors of the target entity based on some relevant relations from all triples associated with the target entity in the KB selected by the LLM. See Appendix \ref{appendix:prompt} for more details.
%     \item \textbf{\texttt{generate(sub-question)}}. Given the sub-question, this action can retrieve related triples via BM25 algorithm \cite{robertson2009} from previous observations (i.e., action execution results), and then prompts the LLM to generate new triples beyond the KB through LLM's parametric knowledge.
%     However, the LLM may frequently return unreliable triples (i.e., inverted head-tail entity pairs and synthesized relationships unsupported by the KB). To address this issue, JCQL prompts the LLM to correct these triples based on the description and schema of relationships (detailed in Appendix \ref{appendix:prompt:supplementary}).
%     \item \textbf{\texttt{finish($e_1,e_2,\dots,e_n$)}}. This action can return a list of entities as final answers, marking the completion of the reasoning process.
% \end{itemize}
 
 Like ReAct \cite{DBLP:conf/iclr/YaoZYDSN023}, we treat the action execution results as observations and the entire reasoning process as a sequence of actions with corresponding observations. Specifically, the agent generates a thought to analyze the current state of the reasoning process before taking action and receives the observation after taking action.
% As shown in Figure \ref{Framework overview}, for each step $i$, JCQL first generates a chain-of-thought trace $\tau_i \in \mathcal{L}$ to analyze the current state, where $\mathcal{L}$ represents the space of language. Subsequently, JCQL selects and executes the action $\alpha_i \in \mathcal{A}$, where $\mathcal{A}=\{\texttt{Search},\texttt{Generate},\texttt{Finish}\}$ is the action space. The result of action is treated as the observation $o_i$. 
Formally, the interaction trajectory at step $i$ can be represented as:
\begin{equation}
    \mathcal{C}_i=(q,E_q,\tau_0,\alpha_0,o_0,\dots,\tau_{i-1},\alpha_{i-1},o_{i-1}),
\end{equation}
where $q$ denotes a question, $E_q$ denotes $q$'s topic entity set, $\alpha_i \in \{\texttt{search},\texttt{generate},\texttt{finish}\}$, $o_i$ denotes the observation, and $\tau_i$ denotes the thought.
% This sequence 
% % $(\tau_0, \alpha_0, o_0, \dots, \tau_{i-1}, \alpha_{i-1}, o_{i-1})$ 
% encapsulates the reasoning traces, actions, and observations from previous steps. 
Based on this interaction trajectory, the generating process for the subsequent thought $\tau_i$ and action $\alpha_i$ can be formulated as follows:
\begin{align}
    & \mathcal{P}(\tau_i|C_i)=\prod_{j=1}^{|\tau_i|}{\mathcal{P}(\tau_i^j|C_i,\tau_{i}^{<j})},\ \ \\
    & \mathcal{P}(\alpha_i|C_i,\tau_i)=\prod_{z=1}^{|\alpha_i|}{\mathcal{P}(\alpha_i^z|C_i,\tau_{i},\alpha_{i}^{<z})}, \label{equal:4}
\end{align}
where $\tau_i^j$ and $|\tau_i|$ are the $j$-th token and the length of $\tau_i$, $\alpha_i^z$ and $|\alpha_i|$ denote the $z$-th token and the length of $\alpha_i$. The LLM repeats this iterative process until it obtains adequate information and outputs final answers in the form of \textbf{\texttt{finish}}.

% \vspace{-1mm}
\subsection{Joint Training for KBC and KBQA} \label{sec:3.3}
% \vspace{-1mm}
% As shown in the left of Figure \ref{Framework overview}, we propose a loop-based joint enhancement framework that synergistically improves both KBC and KBQA via two key mechanisms: (1) KBC $\rightarrow$ KBQA; (2) KBQA $\rightarrow$ KBC. We elaborate them as follows.

\textbf{KBC $\rightarrow$ KBQA.}\label{sec:3.3.1}
To alleviate potential errors caused by the LLM's hallucination in KBQA, we augment the LLM agent by integrating an SLM-trained KBC model as an action within the agent's reasoning process. Specifically, in addition to the three actions \texttt{search}, \texttt{generate}, and \texttt{finish}, we add an additional action denoted as \textbf{\texttt{complete(entity, relation)}}. This action first converts the head entity and the relation into the input format of our SLM-trained KBC model and leverages this KBC model to predict entities. This mechanism creates a tight coupling between KB's explicit structured knowledge and LLM's implicit parametric knowledge to make the answer results more accurate.
Although we constrain the LLM to generate the relation appearing in observations via the design of a specific prompt (detailed in Appendix), it may still generate some relations beyond the scope of the KB. 
% To address this issue, we link the relation generated by the LLM to the most relevant one within the KB by calculating the semantic similarity.
To address this issue, we link the relation generated by the LLM to the most relevant one within the KB based on BM25 scores. 
In practice, a simple approach of determining whether to call the \texttt{complete} action is to set a threshold on the SLM’s decoding probability. However, such threshold is often difficult to obtain. Therefore, we still adopt Formula \ref{equal:4} to calculate the probability of invoking the action \texttt{complete}.

\textbf{KBQA $\rightarrow$ KBC.}\label{sec:3.3.2} 
In order to make KB more complete to assist KBQA, we parse the reasoning paths of KBQA as the supplementary training data to fine-tune the KBC model. As shown in the right of Figure \ref{Framework overview}, given a question-answer pair, we leverage the answers as the supervision to guide the LLM to perform step-by-step reasoning, which makes the LLM agent generate more comprehensive and accurate triples by actively exploring multiple reasoning paths. Specifically, when the reasoning process of a question $q$ is finished, we first compare predicted answers with golden answers and filter the incorrect predicted answers with the corresponding observations from the reasoning process $\mathcal{C}_q$ generated by the LLM agent. 
% After that, the remaining reasoning process needs to be further analyzed to obtain the reasoning paths $P_q$. 
Next, we extract all triples from the reasoning process, including all observations returned from the actions \texttt{search}, \texttt{complete}, and \texttt{generate}. Then, we construct a reasoning subgraph $G_q$ based on the extracted triples, and search for the paths through Breadth-First Search (BFS) from the topic entity to the answer entity in $G_q$. For each entity on the path, we use triples consisting of it and its one-hop neighborhoods to expand the paths. Ultimately, we prompt the LLM to select the triples most relevant to the question based on $q$ and the current paths. The triples selected by the LLM are regarded as the final reasoning paths $P_q$ of KBQA to fine-tune the SLM.

% \textbf{Training Process.} \label{sec:3.3.3}
Based on the above two mechanisms, JCQL's overall training process is depicted in Algorithm \ref{alg:joint training framework}. First, in $\textsc{Pretrain}(\{\langle h,r,t \rangle\}, \mathcal{M}_S)$, JCQL pre-trains a KBC model via the SLM $\mathcal{M}_S$ based on the KBC training set $\{\langle h,r,t\rangle\}$ (introduced in Section \ref{sec:3.1}). Next, in $\textsc{Agent}(q,a_q,e_q,\mathcal{M}_S)$, given a question-answer pair from KBQA's training set, the LLM agent incorporates the pre-trained $\mathcal{M}_S$ to perform reasoning step-by-step to obtain the interaction trajectory $C_q$. Subsequently, in $\textsc{Parse}(C_q,\mathcal{M}_L)$, JCQL leverages the LLM $\mathcal{M}_L$ to extract the reasoning paths $P_q$ based on $C_q$ (introduced in Section \ref{sec:3.3.2}). In $\textsc{IncFineTune}(\{\langle h,r,t\rangle\},P_q,\mathcal{M}_S)$, the triples of $P_q$ are utilized to fine-tune $\mathcal{M}_S$ in real time. Note that each reasoning path contains only a small number of triples, which include those in the KBC's training set used for pre-training (i.e., triples returned from the action \texttt{search}) and supplementary triples provided by $\mathcal{M}_L$ (i.e., triples returned from actions \texttt{complete} and \texttt{generate}). 
To mitigate the issue of catastrophic forgetting, we employ a simple incremental learning technique known as experience replay \cite{DQN} to combine triples in reasoning paths with few triples drawn uniformly at random from replay memory and apply minibatch updates to all these samples to fine-tune $\mathcal{M}_S$. In practice, we store all triples of the KBC's training set in the replay memory.
To reduce training time, we sample only ten triples for each triple in reasoning paths. Thus, by fine-tuning $\mathcal{M}_S$ continuously based on the incremental learning technique, KBC and KBQA can mutually influence each other, in the sense that the KBC model augmented by KBQA's reasoning paths becomes stronger, and the reasoning paths of KBQA incorporating the KBC model are more likely to be high quality. 
% \textcolor{red}{In practice, our method only stores triples from the KBC's training set in the replay memory and samples uniformly at random from the replay memory during updates. 
% This method is in some respects limited since uniform sampling gives equal importance to all triples in replay memory
% .}
To further enhance performance, we can use more advanced incremental learning methods \cite{Incremental_Learning1}, but this is not the focus of our paper. 
% We analyze the time complexity for our training framework. Formally, the time complexity for the whole framework is $O(|Q| \cdot (\omega_{LLM} \cdot D + \omega_{SLM} \cdot (\beta+1) \cdot P))$, where $|Q|$ is the number of training questions, $D$ is the maximum number of \texttt{Thought}, $P$ is the maximum number of triples of reasoning paths, $\beta$ is the number of triples sampled from the KBC training dataset, $\omega_{LLM}$ is the . 

% \begin{algorithm}[t!]
% \caption{JCQL}\label{alg:joint training framework}
% \KwIn{
%   The KB $G$, KBQA's training set $\{Q,A_Q,E_Q\}$, 
%   KBC's training set $\{\langle h,r,t\rangle\in G\}$, 
%   an $SLM$ $\mathcal{M}_S$, an $LLM$ $\mathcal{M}_L$
% }
% \KwOut{
%   The fine-tuned $\mathcal{M}_S$
% }
% % Pre-train the $SLM$
% $\mathcal{M}_S \gets \texttt{Pre\_train}(\{\langle h,r,t\rangle\}, \mathcal{M}_S)$ \;

% \For{\texttt{each} $\{q,a_q,e_q\} \in \{Q,A_Q,E_Q\}$}{
%     % Generate the trajectory for the agent
%     $\texttt{Trajectory}\ C_q \gets \texttt{Agent}(q, a_q, e_q, \mathcal{M}_S)$ \;
    
%     % Parse the trajectory to get paths
%     $\texttt{Paths}\ P_q \gets \texttt{Parse}(C_q, \mathcal{M}_L)$ \;
    
%     % Incrementally fine-tune the model based on the paths
%     $\mathcal{M}_S \gets \texttt{Incremental\_fine\_tuning}(\{\langle h,r,t\rangle\}, P_q, \mathcal{M}_S)$ \;
% }
% \end{algorithm}

% \vspace{-1mm}
\subsection{Inference} \label{sec:3.4}
% \vspace{-1mm}
% Given KBC test triples and the fine-tuned SLM, JCQL predicts $k$ most likely tail entities without requiring scoring all entities in the KB. Specifically, given the query $\langle h, r, ?\rangle$, JCQL first converts it into a one-hop question and combines the question with the corresponding entity description with the context as the input sequence of the fine-tuned SLM (introduced in Section \ref{sec:3.1}). Then, JCQL samples a fixed number of sequences from the decoder of the SLM. Based on the pre-constructed mapping dictionary (e.g., \{Justin Bieber: Q34086\}), the output sequences that do not contain KB entities are filtered. Thus, for each remaining output sequence, we assign a score to the corresponding entity ID based on the probability of decoding the sequence. Simultaneously, for each entity ID that is not output, we assign a negative infinity score. In this way, the top $k$ predicted entities can be returned. Unlike the traditional KBC model, our framework JCQL which is based on the SLM is difficult to assign a score to each entity in the KB. Nevertheless, JCQL still outperforms many SOTA KBC methods, which has been verified in our experiments. 
Given the query $\langle h, r, ?\rangle$ from the KBC's test set and the fine-tuned SLM, JCQL first converts it into the input sequence of the SLM (introduced in Section \ref{sec:3.1}). Then, JCQL samples $R$ sequences from SLM's decoder. Based on the pre-constructed mapping dictionary (e.g., \{Justin Bieber: Q34086\}), the output sequences that do not contain KB entities are filtered. Thus, for each remaining output sequence, we assign a score to the corresponding entity ID based on the probability of decoding the sequence. Simultaneously, we assign a negative infinity score for each entity ID that is not in the output sequences. In this way, the top $k$ predicted entities can be returned. 
Given KBQA's test questions and the fine-tuned SLM, JCQL obtains the answers via the LLM agent without gold answers as the supervision, which is the only difference from the training phase. The results of the action $\texttt{finish}(e_1, e_2, \dots, e_n)$ will serve as final answers. Prompts used here can be found in Appendix.
% \ref{appendix:prompt:framework}.

Formally, the calls to the LLM for the whole inference process are $|Q| \cdot L \cdot N$, where $|Q|$ denotes the number of questions, $L$ denotes the maximum number of \texttt{thought} for each question, and $N$ denotes the maximum number of \texttt{action} calls to the LLM for each question. JCQL first calls the LLM during the action \texttt{search} execution phase to select relevant relations. The second call may occurs during the action \texttt{generate} execution phase, which can generate and correct new triples through the LLM. Thus the value of $N$ is $2$. To reduce the inference time, $L$ is set to $10$.

\begin{center}  % 居中
\begin{minipage}{0.96\columnwidth}  % 限制宽度，防止过宽
\begin{algorithm}[H] % [H] 强制原地显示
\small % 全局小号字体，节省空间
    \caption{JCQL}
    \label{alg:joint training framework}
    % 重定义 Input/Output 样式，使其变为粗体
    \renewcommand{\algorithmicrequire}{\textbf{Input:}}
    \renewcommand{\algorithmicensure}{\textbf{Output:}}
    
    \begin{algorithmic}[1] % [1] 显示行号
        % 【优化】使用 \! 减少集合符号间的空隙，争取在一行内显示更多内容
        \REQUIRE The KB $G$, KBQA's training set $\{Q,A_Q,E_Q\}$, KBC's training set $\{\langle h,r,t\rangle\!\in\!G\}$, an SLM $\mathcal{M}_S$, an LLM $\mathcal{M}_L$
        \ENSURE The fine-tuned $\mathcal{M}_S$
        
        % 逻辑部分
        \STATE $\mathcal{M}_S \leftarrow \textsc{Pretrain}(\{\langle h,r,t\rangle\},\mathcal{M}_S)$
        
        \FOR{$\{q,a_q,e_q\} \in \{Q,A_Q,E_Q\}$}
            % 【优化】使用 \hfill 强制注释右对齐，利用行尾空间
            % \textit{// ...} 模仿 C++ 风格注释，比默认的三角符号更紧凑
            \STATE \textcolor{gray}{\textit{// KBC $\to$ KBQA}}
            \STATE Trajectory $C_q \leftarrow \textsc{Agent}(q,a_q,e_q,\mathcal{M}_S)$

            \STATE \textcolor{gray}{\textit{// KBQA $\to$ KBC}}
            \STATE Paths $P_q \leftarrow \textsc{Parse}(C_q,\mathcal{M}_L)$
            
            \STATE $\mathcal{M}_S \leftarrow \textsc{IncFineTune}(\{\langle h,r,t\rangle\},P_q,\mathcal{M}_S)$
        \ENDFOR
    \end{algorithmic}
\end{algorithm}
\end{minipage}
\end{center}

\renewcommand{\arraystretch}{1.0}
\begin{table*}[t!]
\centering
\setlength{\tabcolsep}{0.8pt}
\resizebox{0.97\textwidth}{!}{%
\begin{tabular}{lccccccccccccccccc}
\toprule
\multirow{2}{*}{\textit{\textbf{Method}}} & \multicolumn{4}{c}{\textit{\textbf{WebQSP (30\% KB)}}} & \multicolumn{4}{c}{\textit{\textbf{WebQSP (50\% KB)}}} & \multicolumn{4}{c}{\textit{\textbf{CWQ (30\% KB)}}} & \multicolumn{4}{c}{\textit{\textbf{CWQ (50\% KB)}}} \\ 
\cmidrule(r){2-5} \cmidrule(r){6-9} \cmidrule(r){10-13} \cmidrule(l){14-17}
                     & \textit{MRR}   & \textit{Hits@1} & \textit{Hits@3} & \textit{Hits@10} & \textit{MRR}   & \textit{Hits@1} & \textit{Hits@3} & \textit{Hits@10} & \textit{MRR}   & \textit{Hits@1} & \textit{Hits@3} & \textit{Hits@10} & \textit{MRR}   & \textit{Hits@1} & \textit{Hits@3} & \textit{Hits@10} \\ 
\midrule
% \multicolumn{17}{c}{\textbf{Structure-based KBC methods}} \\ 
TransE (NeurIPS'13)               & 0.032 & 0.000 & 0.043 & 0.090  & 0.051 & 0.011 & 0.068 & 0.130  & 0.047 & 0.011 & 0.062 & 0.121  & 0.052 & 0.012 & 0.067 & 0.133 \\ 
DistMult (ICLR'15)         & 0.112 & 0.079 & 0.125 & 0.176  & 0.166 & 0.118 & 0.182 & 0.260  & 0.100 & 0.070 & 0.111 & 0.161  & 0.153 & 0.107 & 0.169 & 0.243 \\ 
ComplEx (ICML'16)        & 0.119 & 0.085 & 0.132 & 0.183  & 0.178 & 0.127 & 0.200 & 0.277  & 0.106 & 0.074 & 0.119 & 0.166  & 0.158 & 0.110 & 0.177 & 0.252 \\ 
RotatE (ICLR'19)        & 0.055 & 0.023 & 0.062 & 0.124  & 0.055 & 0.024 & 0.062 & 0.125  & 0.058 & 0.027 & 0.064 & 0.125  & 0.047 & 0.024 & 0.061 & 0.125  \\ 
SimKGC (ACL'22)    & 0.456 & 0.377 & 0.493 & 0.605  & 0.480 & 0.400 & 0.518 & 0.632  & 0.451 & 0.360 & 0.493 & 0.625  & 0.466 & 0.375 & 0.510 & 0.642 \\
% KGT5-context(2023) \cite{kochsiek2023friendly}   & 0.555 & 0.522 & 0.577 & 0.617  & 0.595 & 0.566 & 0.621 & 0.651  & 0.538 & 0.503 & 0.562 & 0.602  & 0.594 & 0.561 & 0.616 & 0.653 \\
KGT5 (ACL'22)            & 0.517 & 0.484 & 0.536 & 0.580  & 0.543 & 0.509 & 0.565 & 0.610  & 0.503 & 0.467 & 0.524 & 0.569  & 0.536 & 0.501 & 0.559 & 0.602  \\ 
BiNet (SIGKDD'22)          & 0.152 & 0.123 & 0.177 & 0.181  & 0.166 & 0.138 & 0.195 & 0.195  & 0.158  & 0.113 & 0.170 & 0.178 & 0.178 & 0.139 & 0.210 & 0.210 \\ 
CSProm-KG (ACL'23)  & 0.416 & 0.360 & 0.448 & 0.518  & 0.432 & 0.375 & 0.466 & 0.538  & 0.400 & 0.343 & 0.433 & 0.506  & 0.421 & 0.363 & 0.454 & 0.529 \\ 
DIFT (ISWC'24)      & 0.515 & 0.457 & 0.540 & 0.625  & 0.536 & 0.474 & 0.563 & 0.649  & 0.504 & 0.438 & 0.530 & 0.634  & 0.521 & 0.454 & 0.547 & 0.651 \\ 
IVR (NeurIPS'24)     & 0.266 & 0.201   & 0.299 & 0.385  & 0.343 & 0.270 & 0.382 & 0.479  & 0.255 & 0.188 & 0.290 & 0.379  & 0.337 & 0.264 & 0.378 & 0.472 \\
GoldE (ICML'24)               & 0.162 & 0.116   & 0.182 & 0.251  & 0.207 & 0.152 & 0.232 & 0.309  & 0.163 & 0.116 & 0.184 & 0.252  & 0.200 & 0.145 & 0.226 & 0.302 \\
RAA-KGC (AAAI'25)              & 0.318 & 0.245   & 0.342 & 0.448  & 0.321 & 0.251 & 0.349 & 0.457  & 0.329 & 0.261 & 0.355 & 0.458  & 0.330 & 0.260 & 0.358 & 0.466 \\
SATKGC (WWW'25)               &0.491 &0.409 &0.531 &0.645 &0.510 &0.427 &0.554 &0.669 &0.476 &0.394 &0.515 &0.630 &0.503 &0.419 &0.546 &0.661\\
% \midrule
% \multicolumn{17}{c}{\textbf{Semantic-based KBC methods}} \\ \hline 
% \midrule
% \multicolumn{17}{c}{\textbf{Joint methods}} \\ \hline 
% \rowcolor{gray!15}

% \rowcolor{gray!15}
JCQL$_{BART-small}$                 
&0.585 &0.553 &0.609 &0.640
&0.609 &0.577 &\underline{0.633} &0.664  
&0.559 &0.526 &0.583 &0.616
&\underline{0.612} &\underline{0.581} &\textbf{0.636} &\underline{0.665}      \\ 
% \rowcolor{gray!15}
JCQL$_{Flan-T5-small}$                 
&\underline{0.590} &\underline{0.559} &\underline{0.612} &\underline{0.646}
&\underline{0.621} &\underline{0.591} &\textbf{0.645} &\underline{0.675}  
&\underline{0.583} &\underline{0.552} &\underline{0.606} &\underline{0.637}
&0.610 &0.578 &0.633 &0.664      \\ 
% \rowcolor{gray!15} 
JCQL$_{T5-small}$                 
    & \textbf{0.603} & \textbf{0.572} & \textbf{0.628} & \textbf{0.652} 
    & \textbf{0.630} & \textbf{0.599} & \textbf{0.652} & \textbf{0.693}   
    & \textbf{0.588} & \textbf{0.556} & \textbf{0.612} & \textbf{0.645}  
    & \textbf{0.614} & \textbf{0.583} & \underline{0.634} & \textbf{0.670}
                                      \\ 
\bottomrule
\end{tabular}%
}
\vspace{-1mm}
\caption{Performance on the KBC task. The best results are indicated in bold. The second best results are underlined.}
\label{Knowledge Graph Completion Performance}
\end{table*}
\renewcommand{\arraystretch}{1.0}

\section{Experimental Study} \label{sec:4}
% \vspace{-1mm}
\subsection{Experiment Settings} \label{sec:4.1}
% \vspace{-1mm}
\textbf{Data Sets.} In the experiments, we utilize two common benchmark data sets, WebQuestionSP (WebQSP) and Complex WebQuestion (CWQ). WebQSP has $4,000$ questions, which consist of both one-hop and two-hop questions. CWQ contains about $30,000$ questions, which include questions requiring composite reasoning and superlative reasoning. Following ToG \cite{sun2023thinkongraph}, we use Wikidata as the background KB. 
% Following previous studies \cite{saxena2020improving, lihui2022binet}, we map the topic entities and the answer entities from Freebase to Wikidata. 
To reduce cost in evaluating our joint task, following \cite{lihui2022binet}, we restrict the KB used in this paper to a subset of Wikidata for the evaluation of all methods by encompassing all triples within two hops of any entity mentioned in WebQSP/CWQ. This scope is small yet enough, covering over 99\% of question answers of WebQSP/CWQ. 
Thus, we can obtain the corresponding background KB for WebQSP and CWQ respectively. For each KB, we randomly sample $20,000$ triples to construct the validation set, $20,000$ triples to construct the test set, and the remaining triples to construct the training set. Following \cite{lihui2022binet}, for each training set, we randomly delete 50\% and 70\% edges as two challenge settings of our work, i.e., 50\% KB and 30\% KB respectively. 
The statistics of these data sets are available in Appendix. We make the data sets and source code used in this paper publicly available for future research\footnote{{https://github.com/ldp2211479/JCQL}}. 
% \footnote{\texttt{https://anonymous.4open.science/r/JCQL/}}
% \footnote{\url{https://anonymous.4open.science/r/JCQL/}}.

% We make the data sets and source code used in this paper publicly available for future research\footnote{\url{https://anonymous.4open.science/r/JCQL/}}. 
% This stochasticity causes different works to have different KBs, making it hard to compare results without re-implementing methods. 

\noindent\textbf{Evaluation Measures.} 
% For KBC, we adopt the evaluation measures mean reciprocal rank (MRR) and Hits@$i$ ($i \in \{1, 3, 10\}$), which are the same as KBC studies \cite{bordes2013, trouillon2016complex}. MRR is calculated by averaging the reciprocal ranks of true tail entities over all test triples. Hits@$i$ evaluates the proportion of true tail entities in the top $i$ predictions. 
% For KBQA, following \cite{sun2023thinkongraph, xu-etal-2024-generate}, we adopt the same evaluation measure Hits@1.
% (the proportion of true answers in the top $1$ predictions) to evaluate all methods. 
For KBC, we adopt the evaluation measures mean reciprocal rank (MRR) and Hits@$i$ ($i \in \{1, 3, 10\}$), which are the same as KBC studies \cite{bordes2013, trouillon2016complex}. For KBQA, following \cite{sun2023thinkongraph, xu-etal-2024-generate}, we adopt the same evaluation measure Hits@1. Please refer to Appendix~\ref{appendix:settings} for more details.

 % For KBQA task, the parameter $k$ is set to $5$.
\noindent\textbf{Setting Details.}
In Section \ref{4.2}, we fix GPT-4o-mini as the LLM and show the performance of JCQL with T5-small, Bart-small \cite{bart}, and Flan-T5-small \cite{flan-t5} as the SLMs. In Section \ref{4.3}, we fix T5-small as the SLM and show the performance of JCQL with Qwen3-30B-A3B \cite{qwen3-embedding}, LLaMA3.1-70B, and GPT-4o-mini as the LLMs. We adopt GPT-4o-mini as the LLM and T5-small as the SLM in other experiments. The maximum size of $\mathcal{T}$, The maximum token length, LLM's temperature parameter, sampling size, and SLM's temperature are set to $20$, $256$, $0.7$, $500$, and $1$ respectively. For the KBQA task, the number of triples returned by the SLM $k$ is set to $5$. JCQL is optimized by using Adafactor with a batch size of 64. For the learning rate, we use the relative learning rate with the warmup initialization method. These hyper-parameters are applied uniformly across all data sets.
Inspired by \cite{sun2023thinkongraph}, we use ReFinED \cite{ayoola2022refinedefficientzeroshotcapableapproach} to perform entity linking for all questions. It is guaranteed that there is at least a path between topic entities and answer entities, which is the same as the previous joint KBC and KBQA study \cite{lihui2022binet}. For questions without topic entities, JCQL answers via CoT \cite{wei2023chainofthoughtpromptingelicitsreasoning}, which is the same setting as the previous KBQA study \cite{sun2023thinkongraph}.

\renewcommand{\arraystretch}{1.0}
\begin{table}[t!]
\centering
\small
\setlength{\tabcolsep}{1.1pt}
\resizebox{0.97\linewidth}{!}{
\begin{tabular}{lccccc}
\toprule
\textit{\textbf{Method}} & \multicolumn{2}{c}{\textit{\textbf{WebQSP}}} & \multicolumn{2}{c}{\textit{\textbf{CWQ}}} \\ 
\midrule
\multicolumn{5}{c}{\textbf{w/o KB}} \\ 
\midrule
GPT-4o-mini                                                                & \multicolumn{2}{c}{0.677} & \multicolumn{2}{c}{0.421} \\ 
CoT (NeurIPS'22)            & \multicolumn{2}{c}{0.646} & \multicolumn{2}{c}{0.446} \\ 
SC (ICLR'23)                & \multicolumn{2}{c}{0.643} & \multicolumn{2}{c}{0.464} \\ 
\midrule
\multicolumn{5}{c}{\textbf{w/ KB}} \\ 
\midrule
& \textit{30\% KB} & \textit{50\% KB} & \textit{30\% KB} & \textit{50\% KB} \\ 
\midrule
DECAF$_{Ans}$ (ICLR'23) & 0.419 & 0.487 & 0.295 & 0.312 \\
KGT5 (ACL'22)      & 0.201 & 0.208 & 0.248 & 0.261 \\
BiNet (SIGKDD'22)      & 0.166 & 0.177 & 0.183 & 0.208 \\ 
KG-CoT (IJCAI'24)         & 0.230 & 0.265 & 0.097 & 0.165 \\
ToG (ICLR'24)                  & 0.457 & 0.706 & 0.360 & 0.328 \\ 
GoG (EMNLP'24)              & 0.684 & 0.693 & \underline{0.476} & \underline{0.477} \\
PoG (NeurIPS'24)              & 0.668 & 0.688 & 0.400 & 0.423 \\
LMP (ACL'25)              & \underline{0.716} & \underline{0.725} & 0.381 & 0.401 \\
PDRR (AAAI'26)              & 0.687 & 0.690 & 0.469 & 0.471 \\
% \midrule
% \multicolumn{5}{c}{\textbf{Joint Methods}} \\ 
% \midrule
% & \textit{30\% KB} & \textit{50\% KB} & \textit{30\% KB} & \textit{50\% KB} \\ \hline
JCQL$_{Qwen3-30B-A3B}$                & 0.668 & 0.676 & 0.422 & 0.427 \\
JCQL$_{LLaMA3.1-70B}$                 & 0.697 & 0.705 & 0.460 & 0.469 \\
JCQL$_{GPT-4o-mini}$                  & \textbf{0.730} & \textbf{0.738} & \textbf{0.502} & \textbf{0.508} \\ 
\bottomrule
\end{tabular}%
}
\vspace{-1mm}
\caption{Performance on the KBQA task in terms of Hits@1. The best results are indicated in bold. The second best results are underlined.}
\label{Knowledge Graph Question Answering Performance}
\end{table}
\renewcommand{\arraystretch}{1}

% \begin{figure*}[t]
%     \centering
%     \includegraphics[width=0.95\linewidth]{image/img/kgqa_for_kgc.pdf}
%     \caption{Ablation study of the effect of KBQA components
% in JQCL}
%     \label{Effect Analysis of Interaction For KGQA}
% \end{figure*}

% \begin{figure*}[t!]
%     \centering
%     \subfigure[]{
%         \label{subfig:(a)}
%         \includegraphics[width=0.235\textwidth]{image/img/kgqa_for_kgc/kgqa_for_kgc_11.pdf}}
%     \subfigure[]{
%        \label{subfig:(b)}
%         \includegraphics[width=0.235\textwidth]{image/img/kgqa_for_kgc/kgqa_for_kgc_12.pdf}}
%     \subfigure[]{
%        \label{subfig:(c)}
%         \includegraphics[width=0.235\textwidth]{image/img/kgqa_for_kgc/kgqa_for_kgc_13.pdf}}
%     \subfigure[]{
%        \label{subfig:(d)}
%         \includegraphics[width=0.235\textwidth]{image/img/kgqa_for_kgc/kgqa_for_kgc_14.pdf}}
%     \subfigure[]{
%        \label{subfig:(e)}
%         \includegraphics[width=0.235\textwidth]{image/img/kgqa_for_kgc/kgqa_for_kgc_21.pdf}}
%     \subfigure[]{
%        \label{subfig:(f)}
%         \includegraphics[width=0.235\textwidth]{image/img/kgqa_for_kgc/kgqa_for_kgc_22.pdf}}
%     \subfigure[]{
%        \label{subfig:(g)}
%         \includegraphics[width=0.235\textwidth]{image/img/kgqa_for_kgc/kgqa_for_kgc_23.pdf}}
%     \subfigure[]{
%        \label{subfig:(h)}
%         \includegraphics[width=0.235\textwidth]{image/img/kgqa_for_kgc/kgqa_for_kgc_24.pdf}}
%     \caption{Ablation study of the effect of the reasoning paths in JCQL.}
%     \vspace{-12pt}
%     \label{Effect Analysis of Interaction For KGC}
% \end{figure*}

\begin{figure*}[t!]
    \centering
    \subfigure[]{
       \label{subfig:(a)}
        \includegraphics[width=0.22\textwidth]{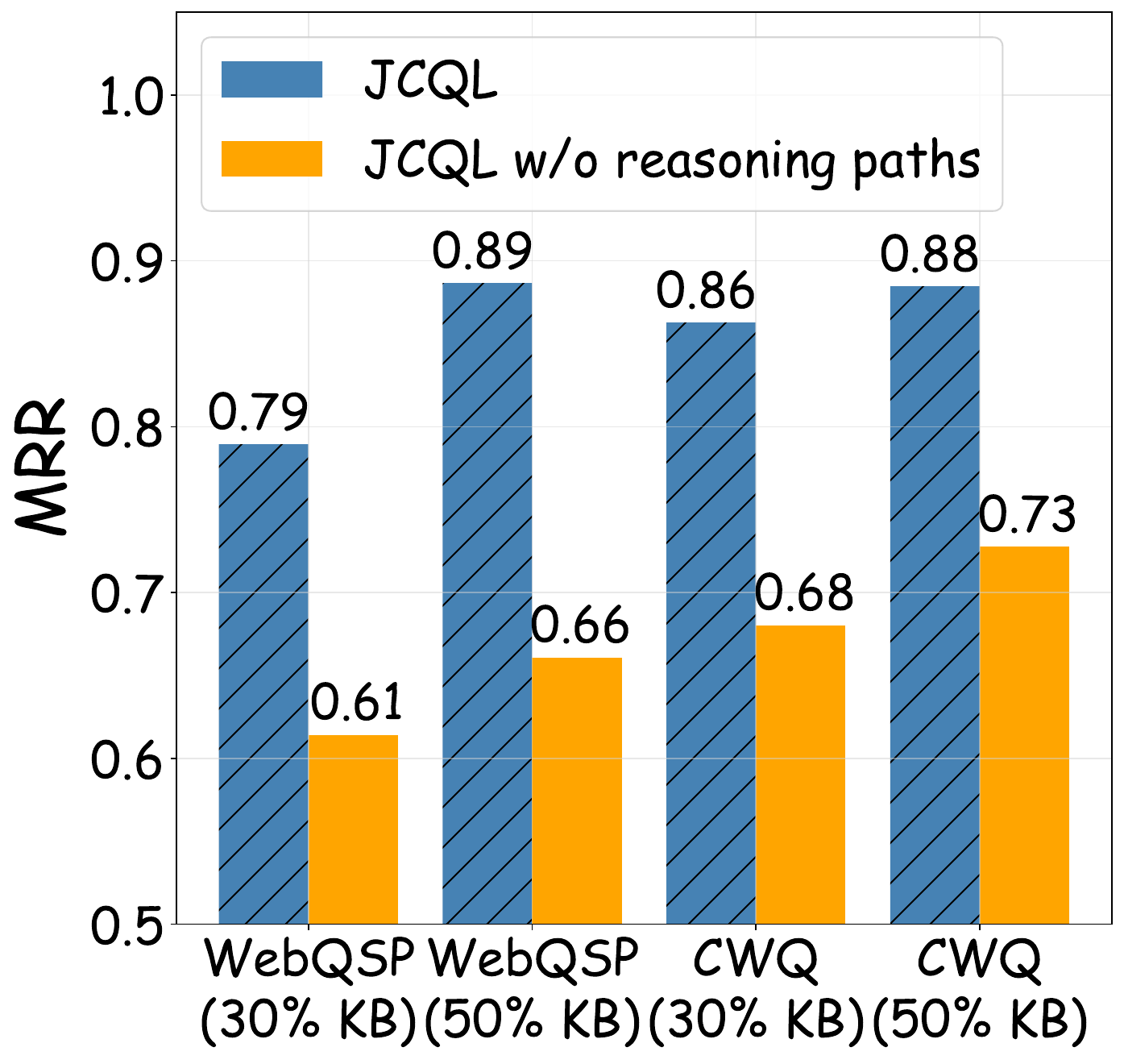}}
    \subfigure[]{
       \label{subfig:(b)}
        \includegraphics[width=0.22\textwidth]{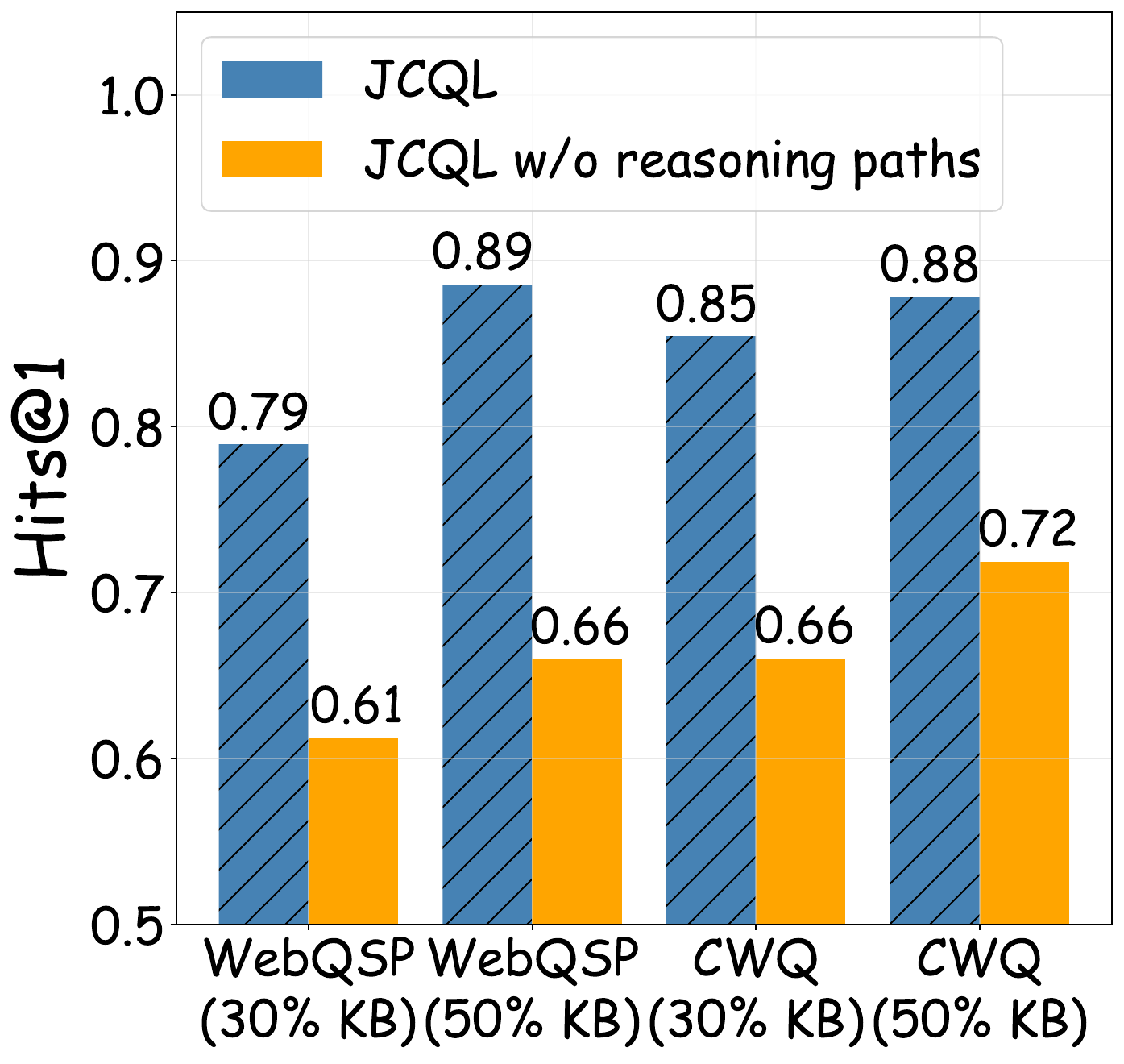}}
    \subfigure[]{
       \label{subfig:(c)}
        \includegraphics[width=0.22\textwidth]{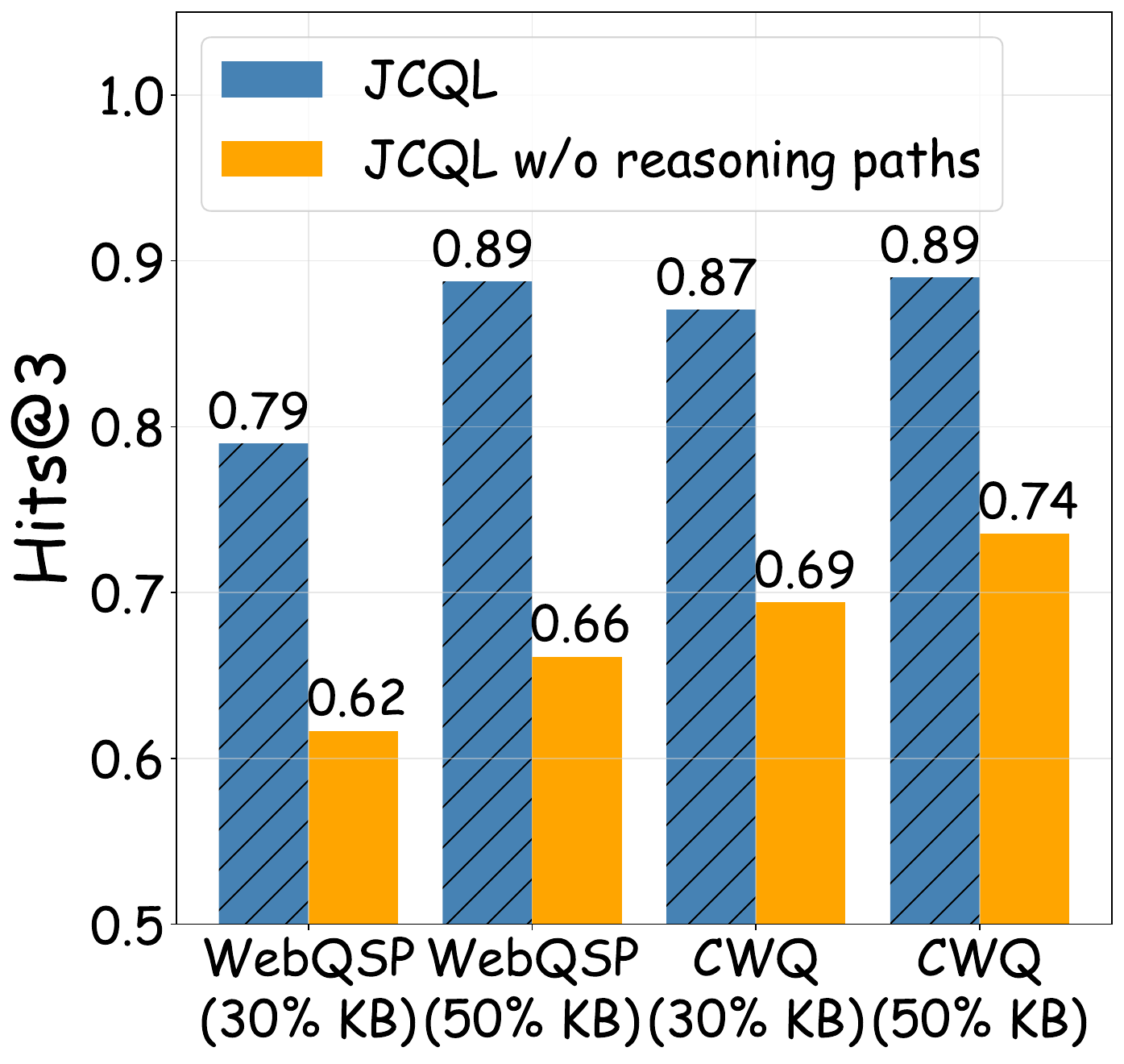}}
    \subfigure[]{
       \label{subfig:(d)}
        \includegraphics[width=0.22\textwidth]{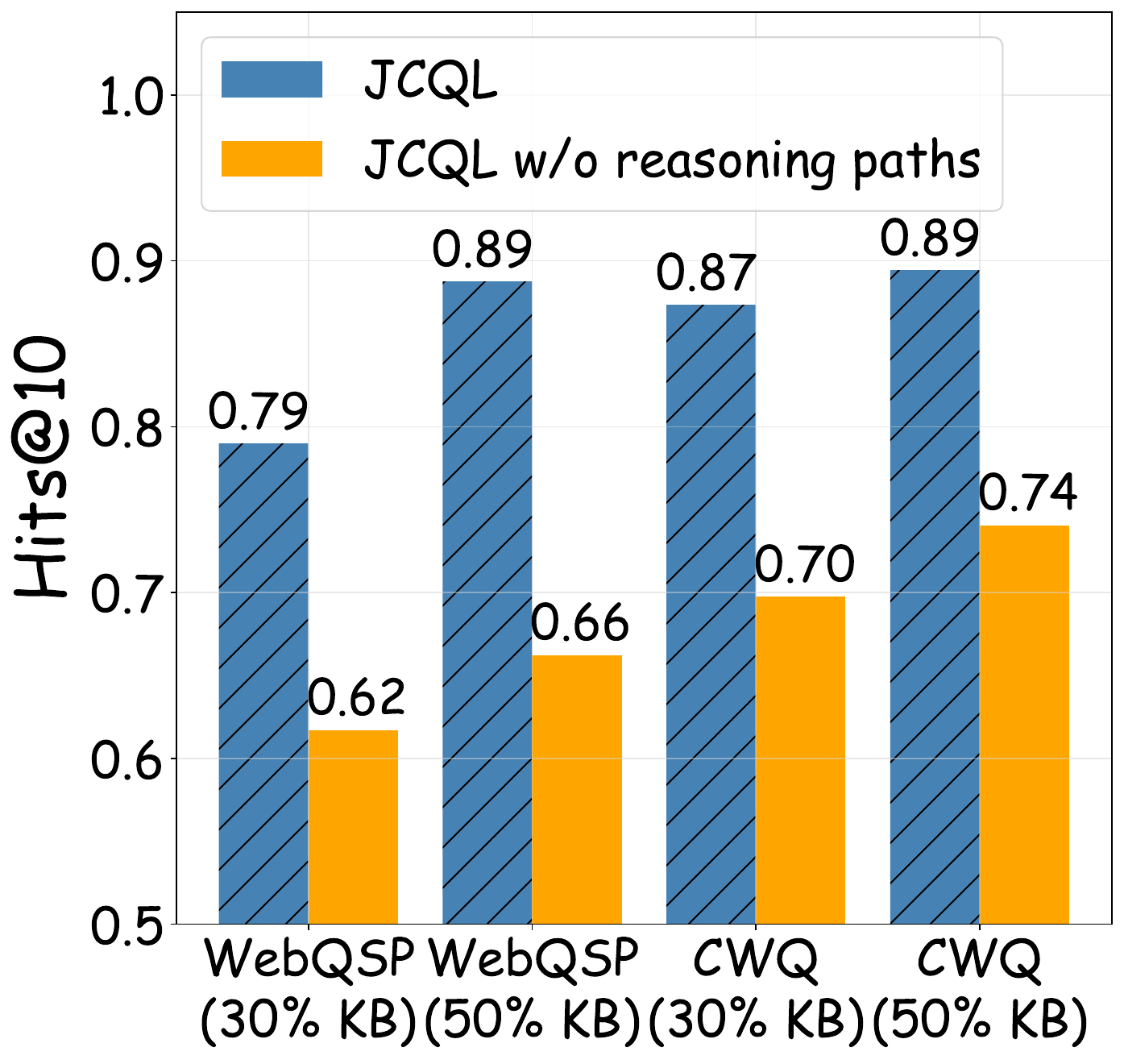}}
    \caption{Ablation study on the LLM's reasoning paths for the task of KBC.}
    % \vspace{-5pt}
    \label{Effect Analysis of Interaction For KGC}
\end{figure*}

% 去掉hits10

% \begin{figure*}[t!]
%     \centering
%     \subfigure[]{
%        \label{subfig:(a)}
%         \includegraphics[width=0.24\textwidth]{image/img/kgqa_for_kgc/kgqa_for_kgc_21.pdf}}
%     \hspace{0.04\textwidth}
%     \subfigure[]{
%        \label{subfig:(b)}
%         \includegraphics[width=0.24\textwidth]{image/img/kgqa_for_kgc/kgqa_for_kgc_22.pdf}}
%     \hspace{0.04\textwidth}
%     \subfigure[]{
%        \label{subfig:(c)}
%         \includegraphics[width=0.24\textwidth]{image/img/kgqa_for_kgc/kgqa_for_kgc_23.pdf}}
%     % \vspace{-4pt}
%     \caption{Ablation study on the LLM's reasoning paths for the task of KBC.}
%     % \vspace{-4pt}
%     \label{Effect Analysis of Interaction For KGC}
% \end{figure*}

\subsection{Knowledge Base Completion} \label{4.2}
For this task, we compare JCQL with $13$ SOTA methods including TransE \cite{bordes2013}, DistMult \cite{yang2015}, ComplEx \cite{trouillon2016complex}, RotatE \cite{sun2018rotate}, IVR \cite{xiao2024knowledge}, GoldE \cite{GoldE}, SimKGC \cite{wang-etal-2022-simkgc}, CSProm-KG \cite{chen-etal-2023-dipping}, DIFT \cite{liu2024finetuninggenerativelargelanguage}, RAA-KGC \cite{raa-kgc}, SATKGC \cite{satkgc}, BiNet \cite{lihui2022binet}, and KGT5 \cite{saxena2022kgt5} on predicting tail entities. Their descriptions are presented in Appendix. Each baseline's performance is reproduced via its open-source solution under its default parameters. 
% For DIFT, we utilize SimKGC as the pre-trained KBC model. For IVR, we use ComplEx as the backbone. For BiNet, we utilize the BFS method to construct relational paths.
From the results on both data sets under different settings (i.e., 50\% KB and 30\% KB) in Table \ref{Knowledge Graph Completion Performance}, it can be seen that JCQL using different SLMs achieves the best performance compared with $13$ baselines in terms of all evaluation measures, demonstrating the effectiveness of JCQL for the KBC task. 

\renewcommand{\arraystretch}{1.0}
\begin{table}[t]
\small
\centering
\setlength{\tabcolsep}{1.0pt}
\begin{tabular}{lccccc}
\toprule
\multirow{2}{*}{\textit{\textbf{Method}}} & \multicolumn{2}{c}{\textit{\textbf{WebQSP}}} & \multicolumn{2}{c}{\textit{\textbf{CWQ}}} \\ \cmidrule(r){2-3} \cmidrule(r){4-5} 
                               & \textit{30\% KB} & \textit{50\% KB} & \textit{30\% KB} & \textit{50\% KB} \\ 
\midrule
JCQL                                 & \textbf{0.730} & \textbf{0.738} & \textbf{0.502} & \textbf{0.508} \\
JCQL w/o  \texttt{generate}                   & 0.706 & 0.722 & 0.482 & 0.490 \\
JCQL w/o  \texttt{complete}                   & 0.691 & 0.696 & 0.473 & 0.475 \\  
\bottomrule
\end{tabular}%
\vspace{-1mm}
\caption{Ablation study on the SLM for the task of KBQA in terms of Hits@1.}
\label{Effect Analysis of Complete Action}
\end{table}
% \end{wraptable}
\renewcommand{\arraystretch}{1}

% \vspace{-1mm}
\subsection{Knowledge Base Question Answering} \label{4.3}
% \vspace{-1mm}
For this task, in addition to BiNet and KGT5, we use other baselines including GPT-4o-mini, Chain-of-Thought (CoT) \cite{wei2023chainofthoughtpromptingelicitsreasoning}, Self-Consistency (SC) \cite{wang2023selfconsistencyimproveschainthought}, KG-CoT \cite{DBLP:conf/ijcai/Zhao0W0024}, ToG \cite{sun2023thinkongraph}, PoG \cite{Plan-on-Graph}, LMP \cite{lmp}, PDRR \cite{pdrr}, GoG \cite{xu-etal-2024-generate} and a variant DecAF$_{Ans}$ (i.e., DecAF only using generated answers) of DecAF that does not need SPARQL queries \cite{decaf}. The descriptions of them are presented in Appendix. For each baseline, the performance is reproduced via its open-source solution under its default parameters. 
% For ToG and GoG, we change the prompt to adapt to Wikidata. For BiNet, we adopt the same setting as introduced in Section \ref{4.2}. 
Note that for all baselines except BiNet and KGT5, we adopt GPT-4o-mini as the LLM. 
% From the results on both data sets under different settings (i.e., $50\%$ KB and $30\%$ KB) in Table \ref{Knowledge Graph Question Answering Performance}, it can be seen that JCQL achieves the best performance compared with $11$ baselines in terms of all evaluation measures, which demonstrates the effectiveness of JCQL for the task of KBQA.
From the results on both data sets under different settings (i.e., $50\%$ KB and $30\%$ KB) in Table \ref{Knowledge Graph Question Answering Performance}, it can be seen that $\text{JCQL}_{GPT-4o-mini}$ achieves the best performance, outperforming all baselines that also employ GPT-4o-mini. Additionally, the variants based on open-source LLMs (i.e., $\text{JCQL}_{LLaMA3.1-70B}$ and $\text{JCQL}_{Qwen3-30B-A3B}$) also exhibit highly competitive results.

\begin{figure*}[t!]
  \centering
  \includegraphics[width=0.98\linewidth]{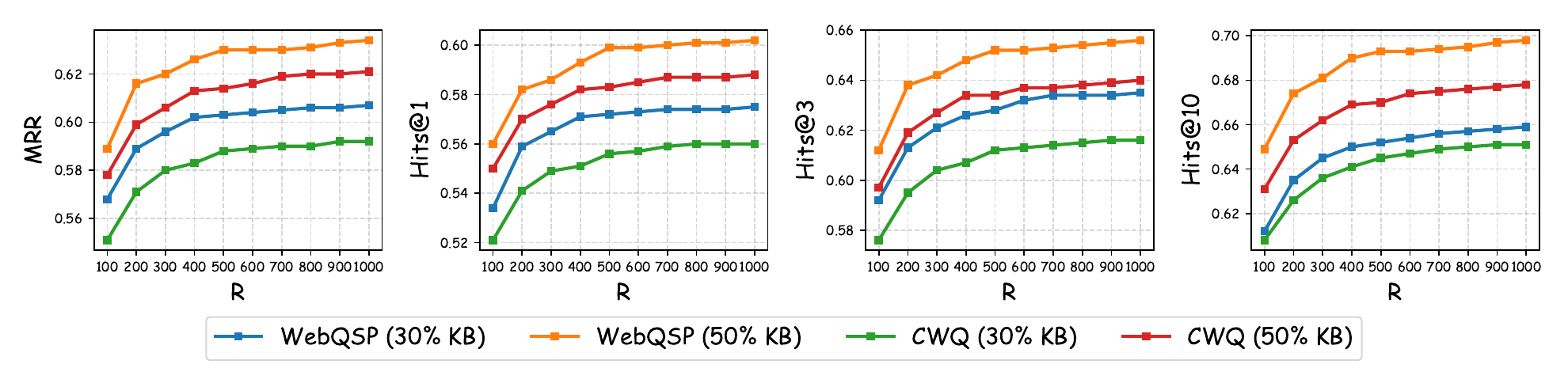}
    % \setlength{\abovecaptionskip}{-0.5mm}
    % \setlength{\belowcaptionskip}{-3mm}
  % \vspace{-1mm}
  \caption{Parameter study over WebQSP and CWQ under different settings.}
  \label{figure:Performance under Different Numbers of the Sequences}
\end{figure*}

% \vspace{-1mm}
\subsection{Ablation Study}
\textbf{Effect Analysis of the LLM's Reasoning Paths in KBC.}
To validate the effectiveness of KBQA in enhancing KBC for JCQL, we conduct a statistical analysis of KBC's test triples and KBQA's reasoning paths in Appendix.
% \ref{appendix:dataset}. 
The analysis results show that the intersection of triples involved in the reasoning paths of the KBQA task and the KBC's test triples is relatively small, which may limit the enhancement of KBQA to KBC. Based on this observation, we first adopt the BFS method to identify the triples that form the shortest path for each KBQA's training question over the complete KB. Then, according to the generated triples of all training questions, we filter the triples that belong to the corresponding KBC training set and use the remaining triples to test the performance of JCQL without reasoning paths. 
% In other words, we extract the missing triples on the reasoning path of each KBQA training question as novel KBC test triples. 
As shown in Figure \ref{Effect Analysis of Interaction For KGC}\subref{subfig:(a)}-\subref{subfig:(d)}, it can be seen that without reasoning paths, the performance of JCQL declines by at least $8$ percentages on both data sets under different settings for the KBC task. Overall, the LLM can improve the SLM’s reasoning ability in KBC, especially for KBC test triples existing in the reasoning paths of KBQA training questions.

\renewcommand{\arraystretch}{1.0}
\begin{table}[t!]
\small
\centering
\setlength{\tabcolsep}{1.0pt}
\resizebox{0.92\linewidth}{!}{
\begin{tabular}{cccccc}
\toprule
\multirow{2}{*}{\textit{\textbf{Data set}}}                  & \multirow{2}{*}{\textit{\textbf{Method}}} & \multicolumn{2}{c}{\textit{\textbf{30\% KB}}} & \multicolumn{2}{c}{\textit{\textbf{50\% KB}}} \\
\cmidrule(r){3-4} \cmidrule(r){5-6}
 & & \textit{LLM call} & \textit{Time (s)}  & \textit{LLM call} & \textit{Time (s)} \\ 
\midrule

\multirow{7}{*}{CWQ}     & ToG   & 18.5   & 33.7 
                                 & 18.2   & 32.7 \\
                         & GoG   & 7.9    & 26.3 
                                 & 7.7    & 26.1 \\
                         & PoG   & 11.7   & 42.6 
                                 & 9.7    & 42.8 \\
                         & LMP   & 11.4   & 59.4 
                                 & 11.3   & 58.8 \\
                         & PDRR  & 8.6    & 49.5  
                                 & 7.6    & 45.8 \\
                         & JCQL-SLM  & 8.2  & 28.4 
                         & 8.0  & 27.2 \\
                         & JCQL  & \textbf{7.3}   & \textbf{25.3} 
                                 & \textbf{7.2}   & \textbf{20.9} \\
                         \midrule
\multirow{7}{*}{WebQSP}  & ToG   & 17.5   & 26.6 
                                 & 15.2   & 23.9 \\
                         & GoG   & 5.9    & 16.8 
                                 & 6.0    & 16.2 \\
                         & PoG   & 6.8    & 31.0 
                                 & 5.4    & 26.2 \\ 
                         & LMP   & 5.4    & 22.1 
                                 & 5.5    & 21.3 \\
                         & PDRR  & 7.5    & 33.1 
                                 & 7.6    & 31.6 \\
                         & JCQL-SLM  & 5.8    & 17.4 
                         & 5.8    & 17.2 \\
                         & JCQL  & \textbf{5.1}   & \textbf{16.6} 
                                 & \textbf{5.2}   & \textbf{15.9} \\ 
\midrule
\end{tabular}
}
\vspace{-1mm}
\caption{Efficiency study on the task of KBQA.}
\label{Effect_Study}
\end{table}
\noindent\textbf{Effect Analysis of the SLM in KBQA.} To verify the importance of the SLM in KBQA, we remove \texttt{complete} and \texttt{generate} from the LLM agent's action set, respectively. \texttt{Generate} prompts the LLM to generate missing triples, while \texttt{complete} generates triples with a fine-tuned SLM.
As shown in Table \ref{Effect Analysis of Complete Action}, we can see that the absence of \texttt{complete} causes a more substantial performance decline in JCQL than the absence of \texttt{generate}, which validates the point that the SLM can indeed alleviate the hallucination to enhance the performance by providing much more accurate tail entities to generate more reliable reasoning paths. 
% Overall, the advantage of the SLM can be utilized to improve the LLM's  shortcomings in KBQA.

% as the additional context of the LLM's reasoning process. 
% It is noted that the action \texttt{Complete} does not affect the inference of the KBC task, so we do not show the result of JCQL without the action  \texttt{Complete} for the KBC task.

\subsection{Efficiency Study}
We study the efficiency of JCQL, JCQL-SLM (i.e., JCQL w/o  \texttt{complete}), and some SOTA baselines (i.e., ToG, PoG, LMP, PDRR, and GoG).
As shown in Table \ref{Effect_Study}, JCQL achieves the best performance in terms of $LLM\  call$ (i.e., the average number of LLM calls for each question) and $Time$ (i.e., inference time) on both data sets under different settings. 
% Specifically, on CWQ, compared with GoG, JCQL reduces the number of LLM calls by approximately $6.9$ percentages. In terms of inference time, JCQL achieves an average of $22.6$ seconds, marking a significant of $13.7$ percentages reduction, which demonstrates the efficiency of JCQL. 
Furthermore, JCQL outperforms JCQL-SLM, demonstrating the effectiveness of the SLM for alleviating the LLM's high computational costs issue.

\subsection{Parameter Study}
In order to comprehensively understand the performance characteristics of JCQL, we conduct the sensitivity analysis to assess how parameter $R$ (the number of sequences from the decoder of the SLM) influences JCQL's performance on the task of KBC. Figure \ref{figure:Performance under Different Numbers of the Sequences} shows the performance of JCQL with parameter $R=\{100,200,300,..., 1000\}$ on WebQSP and CWQ. From the trend shown in Figure \ref{figure:Performance under Different Numbers of the Sequences}, we can see that when the number of sequences increases, the performance improves rapidly at lower values, then grows slowly at high values until it stabilizes on both data sets.

% \vspace{-1mm}

% \vspace{-1mm}
\section{Related Work}
% \vspace{-1mm}
% Three aspects of research are related to our work: (1) KBC; (2) KBQA; (3) joint KBC and KBQA. We will introduce them as follows.
\textbf{KBC} methods often contain two categories of studies: (1) structure-based methods \cite{bordes2013,yang2015,trouillon2016complex,sun2018rotate,xiao2024knowledge,GoldE}; (2) semantic-based methods
% \cite{yao2019kgbertbertknowledgegraph,wei2023kicgpt,liu2024finetuninggenerativelargelanguage,jiang2024kgfit}. 
\cite{wang-etal-2022-simkgc, chen-etal-2023-dipping, saxena2022kgt5, lihui2022binet, liu2024finetuninggenerativelargelanguage}. 
Note that all above methods often rely on offline training with static KBs, resulting in limited scalability, while JCQL uses LLM's reasoning paths to fine-tune an SLM to update the KB dynamically with the incremental learning technique.

% \textbf{KBQA} receives a lot of attention recently. Some studies adopt information retrieval-based methods \cite{sun2023thinkongraph, Plan-on-Graph, xu-etal-2024-generate}.
% CoK \cite{li2024cok} retrieves answers from the KB by parsing questions generated during reasoning into SPARQL queries. 
% Due to the limitations of parsing, ToG \cite{sun2023thinkongraph} and PoG \cite{Plan-on-Graph} attempt to treat the LLM as an agent to interactively explore relation paths step by step in the KB and perform reasoning based on the retrieved paths. KG-CoT \cite{DBLP:conf/ijcai/Zhao0W0024} uses a step-by-step graph reasoning method over the KB to generate high-confidence reasoning paths, enhancing the performance without fine-tuning. GoG \cite{xu-etal-2024-generate} uses the LLM's internal knowledge and KB's external knowledge to reason. However, when relations are lacking, GoG often relies on the LLM's internal knowledge, potentially leading to hallucination. In contrast, to alleviate the hallucination issue, JCQL embeds a fine-tuned SLM-based KBC model as an action into the LLM agent, generating missing triples of the KB via its inherent knowledge and the completion ability of the KBC model.
\textbf{KBQA} receives a lot of attention recently. Some studies adopt information retrieval-based methods \cite{li2024cok, DBLP:conf/ijcai/Zhao0W0024}. ToG \cite{sun2023thinkongraph} and PoG \cite{Plan-on-Graph} attempt to treat the LLM as an agent to interactively explore relation paths step by step in the KB and perform reasoning based on the retrieved paths. GoG \cite{xu-etal-2024-generate} uses the LLM's internal knowledge and the external knowledge in the KB to reason. However, when relations are lacking, GoG often relies on the LLM's internal knowledge, potentially leading to hallucination. In contrast, JCQL embeds a fine-tuned SLM-based KBC model as an action into the LLM agent to generate missing triples, effectively mitigating hallucination issues.

\textbf{Joint KBC and KBQA} methods receive limited attention so far. BiNet \cite{lihui2022binet} converts the question into a relationship path by BERT and jointly deals with KBC and KBQA through a shared embedding space and an answer scoring module.
% Recently, scholars have attempted to integrate knowledge bases into LLMs to provide reliable external knowledge, compensating for insufficient knowledge and hallucination. 
KGT5 \cite{saxena2022kgt5} converts triples into a simple input sequence to pre-train T5 for KBC, and fine-tunes it using question and answer pairs for KBQA. Both methods ignore LLM's reasoning ability, while JCQL combines the LLM and the SLM to integrate the knowledge from the LLM and the KB to solve these two tasks jointly.

\section{Conclusion}
In this paper, we propose a novel framework JCQL, which can resolve KBC and KBQA simultaneously and make the two tasks reinforce each other by combining strengths of the LLM and the SLM. In this framework, LLM's parametric knowledge and KB's structured knowledge can be integrated in an iterative manner. Extensive experiments over two public benchmark data sets have demonstrated the effectiveness of JCQL against many SOTA KBC and KBQA methods.

\section*{Limitations} 
Limitations of JCQL are listed as follows. 
(1) For KBC, the SLM's context is constructed from the one-hop neighborhoods of the head entity. For some challenging queries, this simple context may not provide sufficient information.
(2) For KBQA, JCQL integrates LLM's parametric knowledge and KB's structured knowledge. In future work, we will explore more useful information from the search engine and the integration of multiple KBs to provide more actions for the LLM agent.
(3) The current version of JCQL employs a simple and efficient experience replay strategy. We intend to explore more sophisticated incremental learning techniques to further enhance performance.

\section*{Acknowledgments}
The work was partially supported by the National Key Research and Development Program of China (No. 2024YFF0617702); the National Natural Science Foundation of China (Nos. U22A2025, 62402097, 62232007, and U23A20309); the Joint Funds of the Natural Science Foundation of Liaoning Province (No. 2023-BSBA-132); the 111 Project (No. B16009); the Ant Group Research Program (No. 2025021900003); and the Fundamental Research Funds for the Central Universities (No. N2417007).

\bibliography{custom}

@misc{qwen3-embedding,
      title={Qwen3 Embedding: Advancing Text Embedding and Reranking Through Foundation Models}, 
      author={Yanzhao Zhang and Mingxin Li and Dingkun Long and Xin Zhang and Huan Lin and Baosong Yang and Pengjun Xie and An Yang and Dayiheng Liu and Junyang Lin and Fei Huang and Jingren Zhou},
      year={2025},
      eprint={2506.05176},
      archivePrefix={arXiv},
      primaryClass={cs.CL},
}

@misc{decaf,
      title={DecAF: Joint Decoding of Answers and Logical Forms for Question Answering over Knowledge Bases}, 
      author={Donghan Yu and Sheng Zhang and Patrick Ng and Henghui Zhu and Alexander Hanbo Li and Jun Wang and Yiqun Hu and William Wang and Zhiguo Wang and Bing Xiang},
      year={2023},
      eprint={2210.00063},
      archivePrefix={arXiv},
      primaryClass={cs.CL},
      url={https://arxiv.org/abs/2210.00063}, 
}

@misc{diaglink,
      title={DiagLink: A Dual-User Diagnostic Assistance System by Synergizing Experts with LLMs and Knowledge Graphs}, 
      author={Zihan Zhou and Yinan Liu and Yuyang Xie and Bin Wang and Xiaochun Yang and Zezheng Feng},
      year={2026},
      eprint={2601.20311},
      archivePrefix={arXiv},
      primaryClass={cs.HC},
      url={https://arxiv.org/abs/2601.20311}, 
}

@misc{flan-t5,
      title={Scaling Instruction-Finetuned Language Models}, 
      author={Hyung Won Chung and Le Hou and Shayne Longpre and Barret Zoph and Yi Tay and William Fedus and Yunxuan Li and Xuezhi Wang and Mostafa Dehghani and Siddhartha Brahma and Albert Webson and Shixiang Shane Gu and Zhuyun Dai and Mirac Suzgun and Xinyun Chen and Aakanksha Chowdhery and Alex Castro-Ros and Marie Pellat and Kevin Robinson and Dasha Valter and Sharan Narang and Gaurav Mishra and Adams Yu and Vincent Zhao and Yanping Huang and Andrew Dai and Hongkun Yu and Slav Petrov and Ed H. Chi and Jeff Dean and Jacob Devlin and Adam Roberts and Denny Zhou and Quoc V. Le and Jason Wei},
      year={2022},
      eprint={2210.11416},
      archivePrefix={arXiv},
      primaryClass={cs.LG},
      url={https://arxiv.org/abs/2210.11416}, 
}

@misc{pdrr,
      title={Beyond Chains: Bridging Large Language Models and Knowledge Bases in Complex Question Answering}, 
      author={Yihua Zhu and Qianying Liu and Akiko Aizawa and Hidetoshi Shimodaira},
      year={2025},
      eprint={2505.14099},
      archivePrefix={arXiv},
      primaryClass={cs.CL},
      url={https://arxiv.org/abs/2505.14099}, 
}

@misc{DQN,
      title={Playing Atari with Deep Reinforcement Learning}, 
      author={Volodymyr Mnih and Koray Kavukcuoglu and David Silver and Alex Graves and Ioannis Antonoglou and Daan Wierstra and Martin Riedmiller},
      year={2013},
      eprint={1312.5602},
      archivePrefix={arXiv},
      primaryClass={cs.LG},
      url={https://arxiv.org/abs/1312.5602}, 
}

@article{achiam2023gpt,
  title = {Gpt-4 Technical Report},
  author = {Achiam, Josh and Adler, Steven and Agarwal, Sandhini and Ahmad, Lama and Akkaya, Ilge and Aleman, Florencia Leoni and Almeida, Diogo and Altenschmidt, Janko and Altman, Sam and Anadkat, Shyamal and others},
  year = {2023},
  journal = {arXiv preprint arXiv:2303.08774},
  eprint = {2303.08774},
  archiveprefix = {arXiv}
}

@article{touvron2023llama,
  title = {Llama: {{Open}} and Efficient Foundation Language Models},
  author = {Touvron, Hugo and Lavril, Thibaut and Izacard, Gautier and Martinet, Xavier and Lachaux, Marie-Anne and Lacroix, Timoth{\'e}e and Rozi{\`e}re, Baptiste and Goyal, Naman and Hambro, Eric and Azhar, Faisal and others},
  year = {2023},
  journal = {arXiv preprint arXiv:2302.13971},
  eprint = {2302.13971},
  archiveprefix = {arXiv}
}

@misc{yao2019kgbertbertknowledgegraph,
  title = {{{KG-BERT}}: {{BERT}} for Knowledge Graph Completion},
  author = {Yao, Liang and Mao, Chengsheng and Luo, Yuan},
  year = {2019},
  eprint = {1909.03193},
  primaryclass = {cs.CL},
  archiveprefix = {arXiv}
}

@article{lehmann2015dbpedia,
  title = {Dbpedia--a Large-Scale, Multilingual Knowledge Base Extracted from Wikipedia},
  author = {Lehmann, Jens and Isele, Robert and Jakob, Max and Jentzsch, Anja and Kontokostas, Dimitris and Mendes, Pablo N and Hellmann, Sebastian and Morsey, Mohamed and Van Kleef, Patrick and Auer, S{\"o}ren and others},
  year = {2015},
  journal = {Semantic web},
  volume = {6},
  number = {2},
  pages = {167--195},
}

@article{robertson2009,
  title = {The {{Probabilistic Relevance Framework}}: {{BM25}} and {{Beyond}}},
  author = {Robertson, Stephen and Zaragoza, Hugo},
  year = {2009},
  journal = {Foundations and Trends{\textregistered} in Information Retrieval},
  volume = {3},
  number = {4},
  pages = {333--389},
}

@article{vrandevcic2014wikidata,
  title = {Wikidata: A Free Collaborative Knowledgebase},
  author = {Vrande{\v c}i{\'c}, Denny and Kr{\"o}tzsch, Markus},
  year = {2014},
  journal = {Communications of the ACM},
  volume = {57},
  number = {10},
  pages = {78--85},
}

@article{DBLP:journals/pvldb/FanGZZCCLMDT24,
  author       = {Ju Fan and
                  Zihui Gu and
                  Songyue Zhang and
                  Yuxin Zhang and
                  Zui Chen and
                  Lei Cao and
                  Guoliang Li and
                  Samuel Madden and
                  Xiaoyong Du and
                  Nan Tang},
  title        = {Combining Small Language Models and Large Language Models for Zero-Shot
                  {NL2SQL}},
  journal      = {Proceedings of the VLDB Endowment},
  volume       = {17},
  number       = {11},
  pages        = {2750--2763},
  year         = {2024},
}

@article{T5,
  author       = {Colin Raffel and
                  Noam Shazeer and
                  Adam Roberts and
                  Katherine Lee and
                  Sharan Narang and
                  Michael Matena and
                  Yanqi Zhou and
                  Wei Li and
                  Peter J. Liu},
  title        = {Exploring the Limits of Transfer Learning with a Unified Text-to-Text
                  Transformer},
  journal      = {The Journal of Machine Learning Research},
  volume       = {21},
  pages        = {140:1--140:67},
  year         = {2020},
}

@inproceedings{bart,
    title = "{BART}: Denoising Sequence-to-Sequence Pre-training for Natural Language Generation, Translation, and Comprehension",
    author = "Lewis, Mike  and
      Liu, Yinhan  and
      Goyal, Naman  and
      Ghazvininejad, Marjan  and
      Mohamed, Abdelrahman  and
      Levy, Omer  and
      Stoyanov, Veselin  and
      Zettlemoyer, Luke",
    booktitle = "ACL",
    year = "2020",
    pages = "7871--7880",
}

@inproceedings{raa-kgc,
  author       = {Duanyang Yuan and
                  Sihang Zhou and
                  Xiaoshu Chen and
                  Dong Wang and
                  Ke Liang and
                  Xinwang Liu and
                  Jian Huang},
  title        = {Knowledge Graph Completion with Relation-Aware Anchor Enhancement},
  booktitle    = {AAAI},
  pages        = {15239--15247},
  year         = {2025},
}

@inproceedings{satkgc,
author = {Ko, Youmin and Yang, Hyemin and Kim, Taeuk and Kim, Hyunjoon},
title = {Subgraph-Aware Training of Language Models for Knowledge Graph Completion Using Structure-Aware Contrastive Learning},
year = {2025},
booktitle = {WWW},
pages = {72–85},
}

@inproceedings{lmp,
    title = "Digest the Knowledge: Large Language Models empowered Message Passing for Knowledge Graph Question Answering",
    author = "Wan, Junhong  and
      Yu, Tao  and
      Jiang, Kunyu  and
      Fu, Yao  and
      Jiang, Weihao  and
      Zhu, Jiang",
    booktitle = "{ACL}",
    year = "2025",
    pages = "15426--15442",
}

@inproceedings {Incremental_Learning1,
author = { Jia, Zhifeng and Liu, Hanmo and Li, Haoyang and Chen, Lei },
booktitle = {ICDE},
title = {{ SIT: Selective Incremental Training for Dynamic Knowledge Graph Embedding }},
year = {2025},
pages = {1607-1621},
}

@inproceedings{Plan-on-Graph,
 author = {Chen, Liyi and Tong, Panrong and Jin, Zhongming and Sun, Ying and Ye, Jieping and Xiong, Hui},
 booktitle = {NeurIPS},
 pages = {37665--37691},
 title = {Plan-on-Graph: Self-Correcting Adaptive Planning of Large Language Model on Knowledge Graphs},
 year = {2024}
}

@InProceedings{GoldE,
  title = 	 {Generalizing Knowledge Graph Embedding with Universal Orthogonal Parameterization},
  author =       {Li, Rui and Li, Chaozhuo and Shen, Yanming and Zhang, Zeyu and Chen, Xu},
  booktitle = 	 {ICML},
  pages = 	 {28040--28059},
  year = 	 {2024},
}

@inproceedings{DBLP:conf/iclr/YaoZYDSN023,
  author       = {Shunyu Yao and
                  Jeffrey Zhao and
                  Dian Yu and
                  Nan Du and
                  Izhak Shafran and
                  Karthik R. Narasimhan and
                  Yuan Cao},
  title        = {ReAct: Synergizing Reasoning and Acting in Language Models},
  booktitle    = {ICLR},
  year         = {2023},
  pages        = {1--33}
}

@inproceedings{DBLP:conf/naacl/DevlinCLT19,
  author       = {Jacob Devlin and
                  Ming{-}Wei Chang and
                  Kenton Lee and
                  Kristina Toutanova},
  title        = {{BERT:} Pre-training of Deep Bidirectional Transformers for Language
                  Understanding},
  booktitle    = {NAACL},
  pages        = {4171--4186},
  year         = {2019},
}

@inproceedings{DBLP:conf/ijcai/Zhao0W0024,
  author       = {Ruilin Zhao and
                  Feng Zhao and
                  Long Wang and
                  Xianzhi Wang and
                  Guandong Xu},
  title        = {KG-CoT: Chain-of-Thought Prompting of Large Language Models over Knowledge
                  Graphs for Knowledge-Aware Question Answering},
  booktitle    = {IJCAI},
  pages        = {6642--6650},
  year         = {2024},
}

@inproceedings{li2024cok,
  title = {Chain-of-Knowledge: {{Grounding}} Large Language Models via Dynamic Knowledge Adapting over Heterogeneous Sources},
  booktitle = {ICLR},
  author = {Li, Xingxuan and Zhao, Ruochen and Chia, Yew Ken and Ding, Bosheng and Joty, Shafiq and Poria, Soujanya and Bing, Lidong},
  year = {2024},
  pages={1--23}
}

@inproceedings{sun2023thinkongraph,
  title={Think-on-Graph: Deep and Responsible Reasoning of Large Language Model on Knowledge Graph},
  author={Jiashuo Sun and Chengjin Xu and Lumingyuan Tang and Saizhuo Wang and Chen Lin and Yeyun Gong and Lionel Ni and Heung-Yeung Shum and Jian Guo},
  booktitle={ICLR},
  year={2024},
  pages={1--31}
}

@inproceedings{sun2018rotate,
  title = {{{RotatE}}: {{Knowledge}} Graph Embedding by Relational Rotation in Complex Space},
  booktitle = {ICLR},
  author = {Sun, Zhiqing and Deng, Zhi-Hong and Nie, Jian-Yun and Tang, Jian},
  year = {2019},
  pages = {1649--1667}
}

@inproceedings{wang2023selfconsistencyimproveschainthought,
  author       = {Xuezhi Wang and
                  Jason Wei and
                  Dale Schuurmans and
                  Quoc V. Le and
                  Ed H. Chi and
                  Sharan Narang and
                  Aakanksha Chowdhery and
                  Denny Zhou},
  title        = {Self-Consistency Improves Chain of Thought Reasoning in Language Models},
  booktitle    = {ICLR},
  year         = {2023},
  pages        = {1--24}
}

@inproceedings{yang2015,
  author       = {Bishan Yang and
                  Wen{-}tau Yih and
                  Xiaodong He and
                  Jianfeng Gao and
                  Li Deng},
  title        = {Embedding Entities and Relations for Learning and Inference in Knowledge Bases},
  booktitle    = {ICLR},
  year         = {2015},
  pages        = {1--12}
}

@inproceedings{xiao2024knowledge,
  title = {Knowledge Graph Completion by Intermediate Variables Regularization},
  booktitle = {NeurIPS},
  author = {Xiao, Changyi and Cao, Yixin},
  year = {2024},
  pages = {110218--110245}
}

@inproceedings{10.1145/1376616.1376746,
  title = {Freebase: A Collaboratively Created Graph Database for Structuring Human Knowledge},
  booktitle = {SIGMOD},
  author = {Bollacker, Kurt and Evans, Colin and Paritosh, Praveen and Sturge, Tim and Taylor, Jamie},
  year = {2008},
  pages = {1247--1250},
}

@inproceedings{ayoola2022refinedefficientzeroshotcapableapproach,
    title = "{R}e{F}in{ED}: An Efficient Zero-shot-capable Approach to End-to-End Entity Linking",
    author = "Ayoola, Tom  and
      Tyagi, Shubhi  and
      Fisher, Joseph  and
      Christodoulopoulos, Christos  and
      Pierleoni, Andrea",
    booktitle = "NAACL",
    year = "2022",
    pages = "209--220",
}

@inproceedings{bordes2013,
  title = {Translating {{Embeddings}} for {{Modeling Multi-relational Data}}},
  booktitle = {NeurIPS},
  author = {Bordes, Antoine and Usunier, Nicolas and {Garcia-Duran}, Alberto and Weston, Jason and Yakhnenko, Oksana},
  year = {2013},
  pages = {2787--2795},
}

@inproceedings{brown2020languagemodelsfewshotlearners,
 title = {Language Models Are Few-Shot Learners},
 author = {Brown, Tom and Mann, Benjamin and Ryder, Nick and Subbiah, Melanie and Kaplan, Jared D and Dhariwal, Prafulla and Neelakantan, Arvind and Shyam, Pranav and Sastry, Girish and Askell, Amanda and Agarwal, Sandhini and Herbert-Voss, Ariel and Krueger, Gretchen and Henighan, Tom and Child, Rewon and Ramesh, Aditya and Ziegler, Daniel and Wu, Jeffrey and Winter, Clemens and Hesse, Chris and Chen, Mark and Sigler, Eric and Litwin, Mateusz and Gray, Scott and Chess, Benjamin and Clark, Jack and Berner, Christopher and McCandlish, Sam and Radford, Alec and Sutskever, Ilya and Amodei, Dario},
 booktitle = {NeurIPS},
 pages = {1877--1901},
 year = {2020}
}

@inproceedings{chen-etal-2023-dipping,
  title = {Dipping {{PLMs}} Sauce: {{Bridging}} Structure and Text for Effective Knowledge Graph Completion via Conditional Soft Prompting},
  booktitle = {ACL},
  author = {Chen, Chen and Wang, Yufei and Sun, Aixin and Li, Bing and Lam, Kwok-Yan},
  year = {2023},
  pages = {11489--11503},
}

@inproceedings{lihui2022binet,
  title = {Joint Knowledge Graph Completion and Question Answering},
  booktitle = {SIGKDD},
  author = {Liu, Lihui and Du, Boxin and Xu, Jiejun and Xia, Yinglong and Tong, Hanghang},
  year = {2022},
  pages = {1098--1108},
}

@inproceedings{liu2024finetuninggenerativelargelanguage,
  title={Finetuning generative large language models with discrimination instructions for knowledge graph completion},
  author={Liu, Yang and Tian, Xiaobin and Sun, Zequn and Hu, Wei},
  booktitle={ISWC},
  pages={199--217},
  year={2024},
}

@inproceedings{saxena2022kgt5,
  title = {Sequence-to-Sequence Knowledge Graph Completion and Question Answering},
  booktitle = {ACL},
  author = {Saxena, Apoorv and Kochsiek, Adrian and Gemulla, Rainer},
  year = {2022},
  pages = {2814--2828}
}

@inproceedings{trouillon2016complex,
  title = {Complex Embeddings for Simple Link Prediction},
  booktitle = {ICML},
  author = {Trouillon, Th{\'e}o and Welbl, Johannes and Riedel, Sebastian and Gaussier, {\'E}ric and Bouchard, Guillaume},
  year = {2016},
  pages = {2071--2080}
}

@inproceedings{wang-etal-2022-simkgc,
  title = {{{SimKGC}}: {{Simple}} Contrastive Knowledge Graph Completion with Pre-Trained Language Models},
  booktitle = {ACL},
  author = {Wang, Liang and Zhao, Wei and Wei, Zhuoyu and Liu, Jingming},
  year = {2022},
  pages = {4281--4294},
}

@inproceedings{xu-etal-2024-generate,
  title = {Generate-on-Graph: {{Treat LLM}} as Both Agent and {{KG}} for Incomplete Knowledge Graph Question Answering},
  booktitle = {EMNLP},
  author = {Xu, Yao and He, Shizhu and Chen, Jiabei and Wang, Zihao and Song, Yangqiu and Tong, Hanghang and Liu, Guang and Zhao, Jun and Liu, Kang},
  year = {2024},
  pages = {18410--18430},
}

@inproceedings{wei2023chainofthoughtpromptingelicitsreasoning,
  author       = {Jason Wei and
                  Xuezhi Wang and
                  Dale Schuurmans and
                  Maarten Bosma and
                  Brian Ichter and
                  Fei Xia and
                  Ed H. Chi and
                  Quoc V. Le and
                  Denny Zhou},
  title        = {Chain-of-Thought Prompting Elicits Reasoning in Large Language Models},
  booktitle    = {NeurIPS},
  year         = {2022},
  pages        = {24824--24837},
}

\clearpage
\appendix
\clearpage
\appendix

\section{Statistics of KBs} \label{appendix:dataset}

\paragraph{\textbf{KBC Task.}} The statistics of KBs are shown in Table \ref{Statistics of the 50 KG for the two datasets} and Table \ref{Statistics of the 30 KG for the two datasets}. Note that isolated nodes refer to nodes that do not have any edges connecting to other nodes, and these nodes maintain self-loops (edges pointing to themselves). We retain those samples which have isolated entities (entities without any neighboring nodes). For the entity $e$ that is an isolated node, we add a triple, represented as $\langle e, noop, e \rangle$.
\renewcommand{\arraystretch}{1.2}
\begin{table}[h]
\centering
\setlength{\tabcolsep}{1pt}
\resizebox{0.45\textwidth}{!}{%
    \begin{tabular}{>{\centering\arraybackslash}p{1.4cm}>{\centering\arraybackslash}p{1.4cm}>{\centering\arraybackslash}p{1.4cm}>{\centering\arraybackslash}p{1.6cm}>{\centering\arraybackslash}p{1.4cm}>{\centering\arraybackslash}p{1.4cm}>{\centering\arraybackslash}p{2.3cm}}
    \toprule
    \textit{\textbf{Data set}} & \textit{\textbf{Entities}} & \textit{\textbf{Relations}} & \textit{\textbf{Train}} & \textit{\textbf{Valid}} & \textit{\textbf{Test}} & \textit{\textbf{Isolated nodes}} \\
    \midrule
    WebQSP     & 749,492 & 1,251 & 1,714,822 & 20,000 & 20,000 & 130,179 \\ 
    CWQ        & 672,970 & 1,232 & 1,479,924 & 20,000 & 20,000 & 119,654 \\ 
    \bottomrule
    \end{tabular}%
}
\caption{Statistics of the 50\% KB for the two data sets.}
\label{Statistics of the 50 KG for the two datasets}
\end{table}
\renewcommand{\arraystretch}{1}
\renewcommand{\arraystretch}{1.2}
\begin{table}[h]
\centering
\setlength{\tabcolsep}{1pt}
\resizebox{0.45\textwidth}{!}{%
\begin{tabular}{>{\centering\arraybackslash}p{1.4cm}>{\centering\arraybackslash}p{1.4cm}>{\centering\arraybackslash}p{1.4cm}>{\centering\arraybackslash}p{1.6cm}>{\centering\arraybackslash}p{1.4cm}>{\centering\arraybackslash}p{1.4cm}>{\centering\arraybackslash}p{2.3cm}}
    \toprule
    \textit{\textbf{Data set}} & \textit{\textbf{Entities}} & \textit{\textbf{Relations}} & \textit{\textbf{Train}} & \textit{\textbf{Valid}} & \textit{\textbf{Test}} & \textit{\textbf{Isolated nodes}} \\
    \midrule
    WebQSP  & 749,492 & 1,251 & 1,217,899 & 20,000 & 20,000 & 267,114 \\ 
    CWQ     & 672,970 & 1,232 & 1,060,601 & 20,000 & 20,000 & 244,439 \\ 
    \bottomrule
\end{tabular}%
}
\caption{Statistics of the 30\% KB for the two data sets.}
\label{Statistics of the 30 KG for the two datasets}
\end{table}
\renewcommand{\arraystretch}{1}

\paragraph{\textbf{KBQA Task.}} The statistics of the KBQA data sets used in this paper are shown in Table \ref{Statistics of KGQA datasets.}. Note that QA pairs in the training set without topic entities are removed. It can be seen that simply using the operation edge traverse on the complete KB could achieve nearly 100\% accuracy.
\renewcommand{\arraystretch}{1.2}
\begin{table}[h]
\centering
\setlength{\tabcolsep}{2.5pt}
\resizebox{0.45\textwidth}{!}{%
\begin{tabular}{>{\centering\arraybackslash}p{1.4cm}>{\centering\arraybackslash}p{2.5cm}>{\centering\arraybackslash}p{1.4cm}>{\centering\arraybackslash}p{1.4cm}>{\centering\arraybackslash}p{1.4cm}>{\centering\arraybackslash}p{2cm}}
\toprule
\textit{\textbf{Data set}} & \textit{\textbf{Answer format}} & \textit{\textbf{Train}} & \textit{\textbf{Test}} & \textit{\textbf{Coverage}} & \textit{\textbf{Licence}} \\
\midrule
WebQSP  & Entity/Number & 2777 & 3531 & 0.992 &  CC Licence \\ 
CWQ     & Entity & 23013 & 1639 & 0.997 & -\\ 
\bottomrule
\end{tabular}%
}
\caption{Statistics of KBQA data sets. Coverage refers to the accuracy of subgraph matching.}
\label{Statistics of KGQA datasets.}
\end{table}
\renewcommand{\arraystretch}{1}

\paragraph{\textbf{Reasoning Paths.}} The statistics of reasoning paths are shown in Table \ref{Statistics of Reasoning Paths}. In both KBQA data sets, the inference paths involve less than 1\% of the KBC test triples, indicating that even if the KBQA reasoning path is fully learned, its impact on the KBC test set remains minimal.
\renewcommand{\arraystretch}{1.2}
\begin{table}[h!]
\centering
\setlength{\tabcolsep}{1pt}
\resizebox{0.45\textwidth}{!}{%
    \begin{tabular}{>{\centering\arraybackslash}p{2cm}>{\centering\arraybackslash}p{2.5cm}>{\centering\arraybackslash}p{2.5cm}>{\centering\arraybackslash}p{3.5cm}}
    \toprule
    \textit{\textbf{Data set}} & \textit{\textbf{Number$_{path}$}} & \textit{\textbf{Number$_{test}$}} & \textit{\textbf{Number$_{test}$/Number$_{path}$}} \\
    \midrule
    WebQSP     & 8,623   & 27   & 0.31 \\
    CWQ        & 39,776  & 73   & 0.18 \\
    \bottomrule
    \end{tabular}%
}
\caption{Statistics of the reasoning path. \textit{\textbf{Number$_{path}$}} denotes the number of triples in the inference paths. \textit{\textbf{Number$_{test}$}} denotes the number of triples belonging to inference paths within the KBC test set.}
\label{Statistics of Reasoning Paths}
\end{table}
\renewcommand{\arraystretch}{1}

\section{Supplementary Experiment Settings} \label{appendix:settings}
\textbf{Evaluation Measures.} For KBC, we adopt the evaluation measures mean reciprocal rank (MRR) and Hits@$i$ ($i \in \{1, 3, 10\}$). MRR is calculated by averaging the reciprocal ranks of true tail entities over all test triples. Hits@$i$ evaluates the proportion of true tail entities in the top $i$ predictions. 
For KBQA, we adopt the evaluation measure Hits@1.(the proportion of true answers in the top $1$ predictions) to evaluate all methods. 

\noindent\textbf{KBC task.} For DIFT, we utilize SimKGC as the pre-trained KBC model. For IVR, we use ComplEx as the backbone. For BiNet, we utilize the BFS method to construct relational paths. 

\noindent\textbf{KBQA task.} For ToG and GoG, we change the prompt to adapt to Wikidata. For BiNet, we adopt the same setting as introduced above.

\renewcommand{\arraystretch}{1.3}
\begin{table*}[hbt]
\centering
\setlength{\tabcolsep}{1.2pt}
\resizebox{\textwidth}{!}{
\begin{tabular}{p{0.8cm}p{0.8cm}cccccccccccccccc}
\toprule
\multirow{2}{*}{\textbf{$\mathcal{T}$}} 
& \multirow{2}{*}{\textit{\textbf{\makecell[c]{\textbf{$\mathcal{D}$}}}}} 
& \multicolumn{4}{c}{\textit{\textbf{WebQSP (30\% KB)}}} 
& \multicolumn{4}{c}{\textit{\textbf{WebQSP (50\% KB)}}} 
& \multicolumn{4}{c}{\textit{\textbf{CWQ (30\% KB)}}} 
& \multicolumn{4}{c}{\textit{\textbf{CWQ (50\% KB)}}} \\ 
\cmidrule(r){3-6} \cmidrule(r){7-10} \cmidrule(r){11-14} \cmidrule(l){15-18}

& 
& \textit{MRR}   & \textit{Hits@1} & \textit{Hits@3} & \textit{Hits@10} 
& \textit{MRR}   & \textit{Hits@1} & \textit{Hits@3} & \textit{Hits@10} 
& \textit{MRR}   & \textit{Hits@1} & \textit{Hits@3} & \textit{Hits@10} 
& \textit{MRR}   & \textit{Hits@1} & \textit{Hits@3} & \textit{Hits@10} \\ 
\cmidrule(r){1-18}

\Checkmark   & \Checkmark
    & \textbf{0.603} & \textbf{0.572} & \textbf{0.628} & \textbf{0.652} 
    & \textbf{0.630} & \textbf{0.599} & \textbf{0.652} & \textbf{0.693} 
    & \textbf{0.588} & \textbf{0.556} & \textbf{0.612} & \textbf{0.645} 
    & \textbf{0.614} & \textbf{0.583} & \textbf{0.634} & \textbf{0.670} \\ 

\Checkmark   & \XSolidBrush
&0.566	&0.533	&0.583	&0.626
&0.598	&0.565	&0.620	&0.657
&0.530	&0.496	&0.552	&0.595
&0.582	&0.548	&0.605	&0.643 \\

\XSolidBrush & \Checkmark
    & 0.581 & 0.550 & 0.603 & 0.634    
    & 0.603 & 0.572 & 0.626 & 0.656   
    & 0.570 & 0.539 & 0.590 & 0.625  
    & 0.598 & 0.568 & 0.620 & 0.650   \\

\XSolidBrush & \XSolidBrush
    & 0.517 & 0.485 & 0.537 & 0.581    
    & 0.544 & 0.509 & 0.565 & 0.610 
    & 0.503 & 0.467 & 0.524 & 0.569  
    & 0.536 & 0.501 & 0.559 & 0.602 \\
\bottomrule
\end{tabular}%
}
\caption{Ablation study of the effect of the input sequence form of the SLM.}
\label{Effect Analysis of the information for Joint Fine-Tuning T5}
\end{table*}
\renewcommand{\arraystretch}{1}

\section{Baselines} \label{appendix:baseline}
\subsection{KBC Methods} \label{appendix:baseline:kbc}
For this task, we compare JCQL with eleven SOTA approaches as follows: 
\begin{itemize}[left=0pt]
\item TransE \cite{bordes2013} computes the candidate tail entity' score by representing relationships as vector translations.
\item DistMult \cite{yang2015} uses a bilinear scoring function with a diagonal relationship matrix to calculate scores of candidate tail entities.
\item ComplEx \cite{trouillon2016complex} extends DistMult by embedding entities and relations in a complex space.
\item RotatE \cite{sun2018rotate} obtains the score of tail entities by defining the relationship as rotations to capture a wider range of relational patterns.
\item IVR \cite{xiao2024knowledge} represents entities and relationships as embedding matrices and computes scores using a tensor decomposition model.
\item GoldE \cite{GoldE} computes the score of tail entities by applying hyperbolic orthogonal transformations to head entities and calculating the inner product with tail entities to capture hierarchical and geometric structures in knowledge graphs.

\item SimKGC \cite{wang-etal-2022-simkgc} leverages contrastive learning to optimize query embeddings derived from BERT, and computes the scores of candidate tail entities using the cosine similarity.
\item CSProm-KG \cite{chen-etal-2023-dipping} uses the embedding of entities and relationships to generate conditional soft prompts, which are then entered into the frozen SLM to predict the tail entity.
\item DIFT \cite{liu2024finetuninggenerativelargelanguage} predicts candidate tail entity by a pre-trained KBC model and refines the selection process with LLaMA.
\item SATKGC \cite{satkgc} proposes a subgraph-aware training framework that incorporates structural inductive biases into SLMs by utilizing random-walk based subgraph sampling as mini-batches and employing contrastive learning to prioritize topologically hard negatives based on shortest path distances.
\item RAA-KGC \cite{raa-kgc} proposes a relation-aware anchor enhancement strategy that utilizes neighbors of the head entity sharing the same relation as context.
\item BiNet \cite{lihui2022binet} converts multi-hop questions into relational paths via an encoder-decoder structure and utilizes a shared embedding space with an answer scoring module for joint optimization of KBC and KBQA.
\item KGT5 \cite{saxena2022kgt5} integrates KBC and KBQA tasks into a unified framework by converting the query and the question into a simple input sequence form and fine-tuning the model T5.
\end{itemize}

\begin{figure*}[t!]
    \centering
    \subfigure[]{
       \label{training_time:subfig:(a)}
        \includegraphics[width=0.4\textwidth]{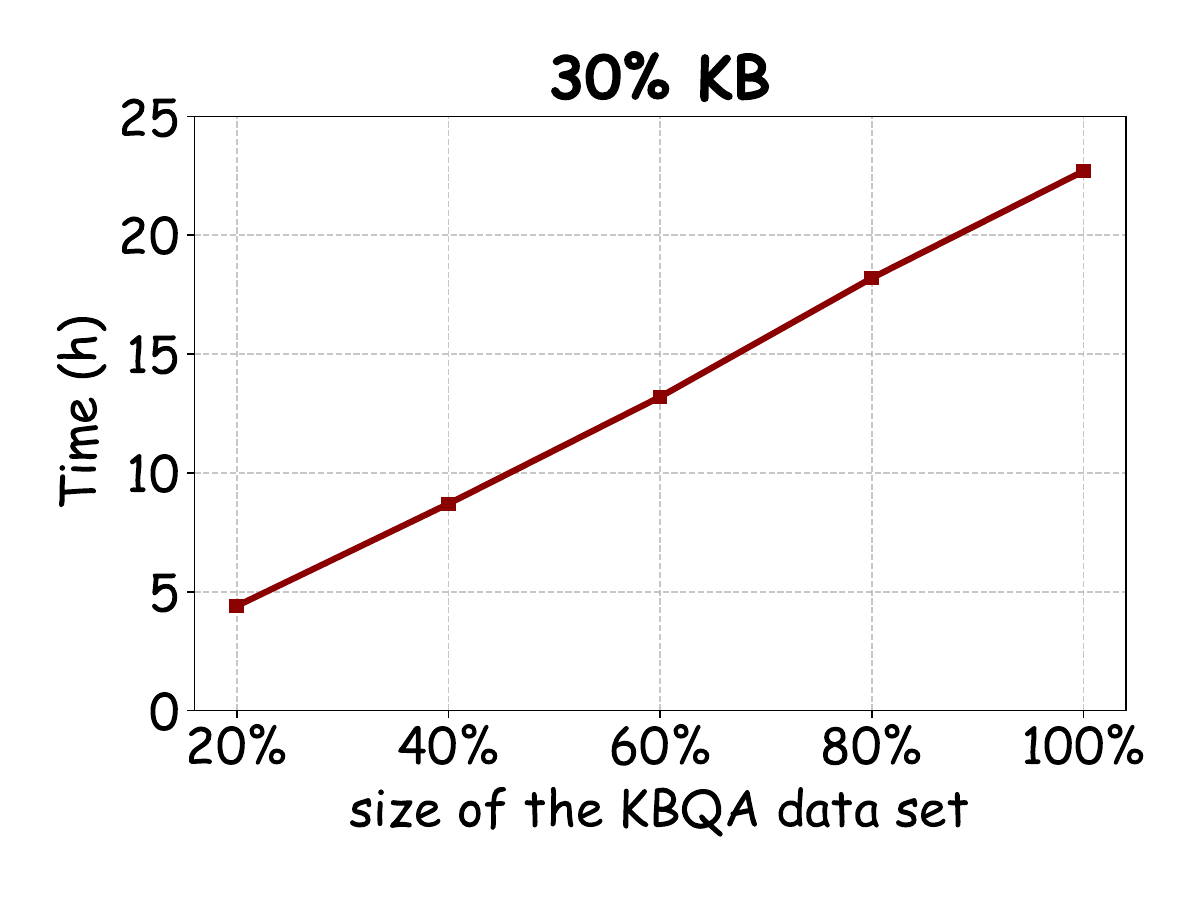}}
    \subfigure[]{
       \label{training_time:subfig:(b)}
        \includegraphics[width=0.4\textwidth]{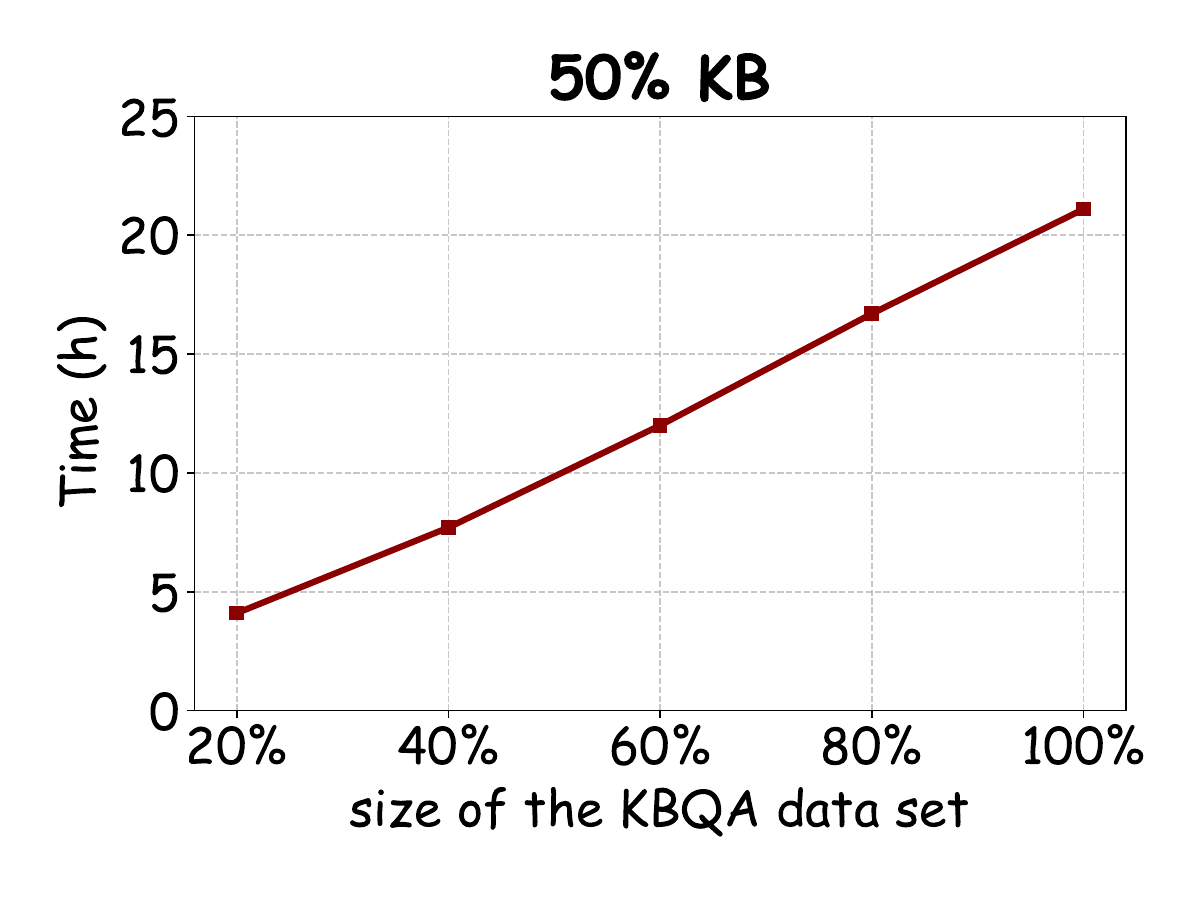}}
    % \vspace{-3pt}
    \caption{Efficiency study of the training process on WebQSP.}
    \label{training time}
\end{figure*}
\begin{figure*}[t!]
  \centering
  \includegraphics[width=0.95\linewidth]{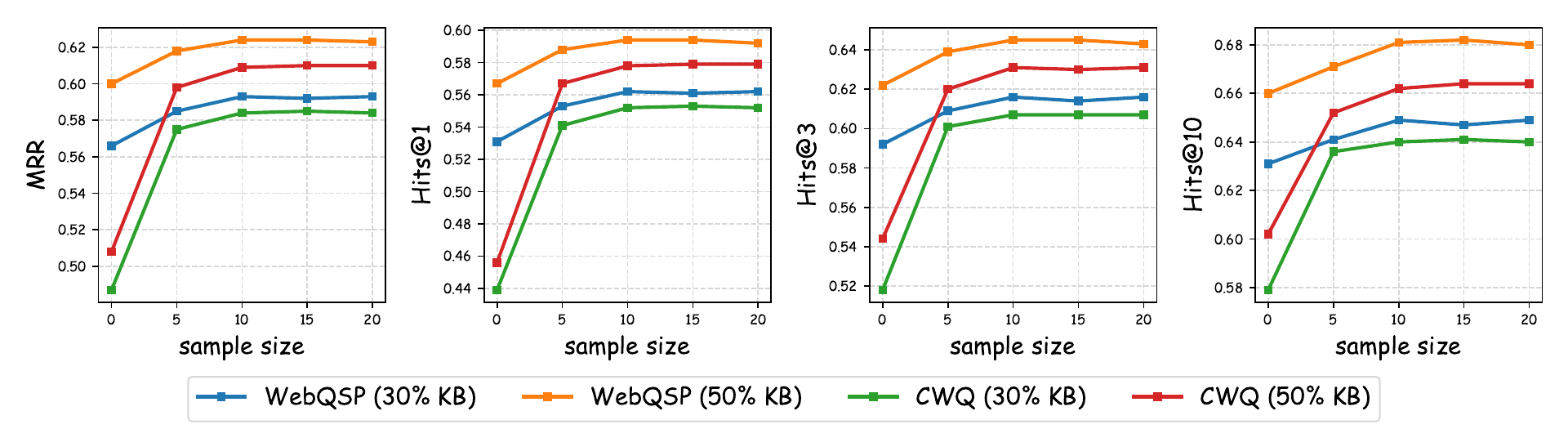}
  \caption{Performance of JCQL with different sample numbers in experience replay.}
  \label{figure:Performance under Different Sample Numbers}
\end{figure*}

\subsection{KBQA Methods} \label{appendix:baseline:kbqa}
For this task, in addition to BiNet and KGT5, we use other baselines as follows:
\begin{itemize}[left=0pt]
\item GPT-4o-mini is adopted under the standard prompting\cite{brown2020languagemodelsfewshotlearners} without requiring the KB.
\item Chain-of-Thought (CoT) prompt \cite{wei2023chainofthoughtpromptingelicitsreasoning} solves questions through step-by-step reasoning without requiring the KB.
\item Self-Consistency (SC) \cite{wang2023selfconsistencyimproveschainthought} can generate multiple candidate answers via CoT and select the most consistent answer.
\item KG-CoT \cite{DBLP:conf/ijcai/Zhao0W0024} leverages a step-by-step graph reasoning model to perform reasoning over KBs and generates chains of knowledge with high confidence for LLMs based on the reasoning paths.
\item ToG \cite{sun2023thinkongraph} explores relation paths within KBs through iterative interactions with the LLM and let the LLM derive final answers based on the retrieved paths.
\item PoG \cite{Plan-on-Graph} adaptively explores and self-corrects reasoning paths to efficiently answer questions by leveraging guidance, memory, and reflection.
\item GoG \cite{xu-etal-2024-generate} retrieves triples from the KB and generates novel triples through the LLM to answer complex questions in the incomplete KB.
\item LMP \cite{lmp} adapts the message passing paradigm to the textual domain by iteratively condensing neighborhoods into hierarchical semantic facts, enabling LLMs to digest structural knowledge for effective reasoning.
\item PDRR \cite{pdrr} employs a semantic parsing-inspired framework to predict question types and decompose them into structured triples, enabling flexible planning for both chain and parallel reasoning over KGs.
\end{itemize}

\section{Supplementary Experimental Results} \label{appendix:supplementary_results}
% \subsection{Knowledge Base Completion} To gain deeper insight into the performance of JCQL in KBC, we conduct a comparative analysis of all baseline models with the Hits@10 metric. As shown in Table \ref{Knowledge Graph Completion Performance hits10}, JCQL also achieving the best performance in terms of Hits@10. % 为了进一步探究JCQL在KBC上的贡献，我们测试了所有baseline在Hits@10的性能。如图\ref{tables/KGC_compare_hits10}所示，JCQL在Hits@10上仍然达到了最佳性能。

\subsection{Effect Analysis of the Input Sequence Form of the SLM}
To verify the effectiveness of two kinds of information (i.e., entity description \textbf{$\mathcal{D}$} and context \textbf{$\mathcal{T}$}) introduced in Section \ref{sec:3.3}, we remove them from JCQL. As shown in Table \ref{Effect Analysis of the information for Joint Fine-Tuning T5}, without \textbf{$\mathcal{D}$} and \textbf{$\mathcal{T}$}, the performance of JCQL declines for KBC task on both data sets under different settings, which demonstrates the effectiveness of them. 

% \subsection{Effect Analysis of Different Numbers of Completed Triples}
% We investigate how the number of completed triples influence JCQL's performance. As depicted in Figure \ref{num_complete_triples}, JCQL performance generally improves with more triples, peaking before declining. This decline is likely due to the noise and irrelevant knowledge introduced by excessive complementation. Crucially, complementing the knowledge base with any number of predicted triples enhances the reasoning capacity for the KBQA task compared to the absence of complementation (i.e., $k=0$).

\subsection{Efficiency Study of Training Process}
In this subsection, we study the efficiency of JCQL's training process under different settings. Figure \ref{training_time:subfig:(a)} and Figure \ref{training_time:subfig:(b)} plot the total running time on WebQSP under different settings i.e., 30\% and 50\% of the KB, respectively. From the results, we can see that the total running time is approximately linear to the number of questions in the training data set, which is consistent with calls to the LLM described in Section \ref{sec:3.4}. Note that the running time of $\textsc{Parse}(C_q,\mathcal{M}_L)$ and $\textsc{IncFineTune}(\{\langle h,r,t\rangle\},P_q,\mathcal{M}_S)$ is negligible compared with $\textsc{Agent}(q,a_q,e_q,\mathcal{M}_S)$. Therefore, the training time mainly depends on $\textsc{Agent}(q,a_q,e_q,\mathcal{M}_S)$.

\subsection{Performance of JCQL with different sample size in experience replay.} To explore the influence of the sample size in experience replay on JCQL’s performance, we conduct experiments with the size set to $\{0, 5, 10, 15, 20\}$. As shown in Figure \ref{figure:Performance under Different Sample Numbers}, JCQL’s performance improves with the sample size. 
This suggests that our experience replay strategy can effectively mitigate catastrophic forgetting in SLM. It also demonstrates that performance growth diminishes when the sample size exceeds 10. To balance performance and computational cost, we set the sample size to 10.

% \begin{figure}[ht]
%     \centering
%     \includegraphics[width=0.75\linewidth]{image/img/num relevant triples in Complete Action.pdf}
%     \caption{Performance under different number of the completed triples}
%     \label{num_complete_triples}
% \end{figure}

% \begin{figure*}[t!]
%     \centering
%     \subfigure[WebQSP]{
%        \label{completed:subfig:(a)}
%         \includegraphics[width=0.35\textwidth]{image/img/num relevant triples in Complete Action left.pdf}}
%     \subfigure[CWQ]{
%        \label{completed:subfig:(b)}
%         \includegraphics[width=0.35\textwidth]{image/img/num relevant triples in Complete Action right.pdf}}
%     \vspace{-5pt}
%     \caption{Performance under different number of the completed triples}
%     \label{num_complete_triples}
% \end{figure*}

% \begin{figure*}[t!]
%     \centering
%     \includegraphics[width=1\linewidth]{image/img/num relevant triples in Complete Action.pdf}
%     \vspace{-9pt}
%     \caption{Performance under different number of completed triples.}
%     \label{num_complete_triples}
% \end{figure*}

\section{Prompt List} \label{appendix:prompt}
% \subsection{Prompts in Training Phase}\label{appendix:prompt:training phase}
% The following shows the prompts used in JCQL including Knowledge Extraction Prompt, KBQA Inference Selecting Entity and Template Consistency Prompt.

\subsection{Prompts for JCQL}\label{appendix:prompt:framework}
As shown in Prompt \ref{prompts for instruction used in knowledge extraction}, the prompt is designed to extract the reasoning process of KBQA. 
% \textcolor[rgb]{0,0.4392,0.7529}{The blue font} indicates the retrieved answers, \textcolor[rgb]{0.7529,0,0}{the red font} indicates \texttt{Generate Action} and \textcolor[rgb]{0.4549,0.7137,0.0078}{the green font} indicates \texttt{Complete Action}.
\begin{tcolorbox}[prompt=
{\begin{prompt} \label{prompts for instruction used in knowledge extraction}The Prompt for the Training Phase of JCQL 
\end{prompt}}]
$\bullet$ Instruction: \\ \\
Solve a question answering task with interleaving Thought, Action, Observation steps. Thought can reason about the current situation, and Action can be four types: \\
\textbf{(1) Search[entity1 | entity2 | ...],} which searches the exact entities on Wikidata and returns their one-hop subgraphs. You should extract all concrete entities appeared in your last thought without redundant words, and you should always select entities from topic entities in the first search. \\
\textbf{(2) Complete[head entity | relation],} which completes tail entities based on the head entity and relationships. \\ 
\textbf{(3) Generate[subquestion],} which generates some new triples related to your last thought. The subquestion should be a one-hop question. \\
\textbf{(4) Finish[answer1 | answer2 | ...],} which summarizes the answers from the triples and returns the true answers. You should reason about the answer based on the known triples. Firstly, you need to summarize all possible answers appeared in the known triples. Secondly, you should compare them with the reference answers. Finally, you should return all the true answers. If you don't know the answer, please return Finish[unknown].\\
You should generate each step without redundant words. \\

$\bullet$ Example: \\ \\
\textbf{Question:} where did woodrow wilson go to school? \\
\textbf{Topic Entity:} [Woodrow Wilson] \\
\textbf{Answer:} Davidson College | University of Virginia School of Law | Princeton University | Johns Hopkins University \\
\textbf{Thought 1:} The task requires identifying educational institutions attended by Woodrow Wilson. Based on the question, I will focus on gathering educational institutions linked to him. \\
\textbf{Action 1:} Search[Woodrow Wilson] \\
\textbf{Observation 1:} Woodrow Wilson, educated at, \textcolor[rgb]{0,0.4392,0.7529}{Davidson College} \\
Woodrow Wilson, employer, Bryn Mawr College \\
Woodrow Wilson, educated at, University of Virginia \\
\textbf{Thought 2:} While Davidson College and the University of Virginia are identified, I suspect there are more institutions. I will use the "Complete" action with "Woodrow Wilson" and the relation "educated at" to find additional schools. \\
\textbf{Action 2:} \textcolor[rgb]{0.4549,0.7137,0.0078}{Complete[Woodrow Wilson | educated at]} \\
\textbf{Observation 2:} Woodrow Wilson, educated at, \textcolor[rgb]{0,0.4392,0.7529}{Princeton University} \\
\textbf{Thought 3:} The result of Complete Action shows that Princeton University is also the school attended by Woodrow Wilson. I need to conduct further analysis to determine if there are other possible institutions. \\
\textbf{Action 3:} \textcolor[rgb]{0.7529,0,0}{Generate[which additional educational institutions are linked to Woodrow Wilson?]} \\
\textbf{Observation 3:} Woodrow Wilson, educated at, \textcolor[rgb]{0,0.4392,0.7529}{Johns Hopkins University} \\
\textbf{Thought 4:} Firstly, I have gathered four educational institutions associated with Woodrow Wilson: Davidson College, University of Virginia, Princeton University, and Johns Hopkins University. Secondly, I compare them completely with the reference answers. The University of Virginia School of Law is missing. \\
\textbf{Action 4:} Finish[Davidson College | University of Virginia | Princeton University | Johns Hopkins University] \\
\textbf{Observation 4:} Davidson College | Princeton University | Johns Hopkins University
\end{tcolorbox}
% \input{prompts/prompts for example used in knowledge extraction}

% \subsection{Prompts in Inference Phase}\label{appendix:prompt:inference phase}
As shown in Prompt \ref{prompts for instruction used in KGQA inference.}, the prompt is designed to executive reasoning for KBQA via the agent. \textcolor[rgb]{0,0.4392,0.7529}{The blue font} denotes the retrieved answers, \textcolor[rgb]{0.7529,0,0}{the red font} denotes the action \texttt{generate}, and \textcolor[rgb]{0.4549,0.7137,0.0078}{the green font} denotes the action \texttt{complete}.

\begin{tcolorbox}[prompt=
{\begin{prompt} \label{prompts for instruction used in KGQA inference.} The Prompt for the Inference Phase of JCQL
% Prompts for instruction used in the inference phase for KBQA task.
\end{prompt}}]

$\bullet$ Instruction: \\ \\
Solve a question answering task with interleaving Thought, Action, Observation steps. Thought can reason about the current situation, and Action can be four types:  \\
\textbf{(1) Search[entity1 | entity2 | ...],} which searches the exact entities on Wikidata and returns their one-hop subgraphs. You should extract all concrete entities appeared in your last thought without redundant words, and you should always select entities from topic entities in the first search.  \\
\textbf{(2) Complete[head entity | relation],} which completes tail entities based on the head entity and relationships. \\ 
\textbf{(3) Generate[subquestion],} which generates some new triples related to your last thought. The subquestion should be a one-hop question.  \\
\textbf{(4) Finish[answer1 | answer2 | ...],} which summarizes the answers based on all the known triples. You should reason about the answer based on the known triples. If you don't know the answer, please return Finish[unknown]. \\
You should generate each step without redundant words. \\

$\bullet$ Example: \\ \\
\textbf{Question:} Which one of Shakespeare's parents worked as a business person? \\
\textbf{Topic Entity:} [William Shakespeare] \\
\textbf{Thought 1:} I need to find Shakespeare's parents firstly. \\
\textbf{Action 1:} Search[William Shakespeare] \\
Observation 1: Judith Quiney, father, William Shakespeare \\
\textbf{Thought 2:} The search results did not provide specific information about William Shakespeare's parents or their occupations. Therefore, I need to identify Shakespeare's father. \\
\textbf{Action 2:} Generate[who are William Shakespeare's father?] \\
\textbf{Observation 2:} William Shakespeare, father, John Shakespeare \\
\textbf{Thought 3:} I could find the father of William Shakespeare is John Shakespeare. I need to identify Shakespeare's mother then. \\
\textbf{Action 3:} \textcolor[rgb]{0.7529,0,0}{Generate[who are William Shakespeare's mother?]} \\
Observation 3: William Shakespeare, mother, Mary Shakespeare \\
\textbf{Thought 4:} According to the generate results, I can find the father of William Shakespeare is John Shakespeare and the mother is Mary Shakespeare. So next I will search for some information about their occupations. \\
\textbf{Action 4:} Search[John Shakespeare | Mary Shakespeare] \\
\textbf{Observation 4:} John Shakespeare, child, Joan Shakespeare \\
Mary Shakespeare, spouse, John Shakespeare \\
Mary Shakespeare, occupation, writer \\
\textbf{Thought 5:} I find the occupation of Mary Shakespeare is writer, but the search results did not provide specific information about John Shakespeare's occupations. Therefore, I need to identify his business-related activities. \\
\textbf{Action 5:} \textcolor[rgb]{0.4549,0.7137,0.0078}{Complete[John Shakespeare | occupation]} \\
Observation 5: \textcolor[rgb]{0,0.4392,0.7529}{John Shakespeare}, occupation, merchant \\
John Shakespeare, occupation, politician \\
\textbf{Thought 6:} John Shakespeare is identified as William Shakespeare's father with occupations as a glove maker and a local government official. This connects him to the business realm. Therefore, I can summarize the answers. \\
\textbf{Action 6:} Finish[John Shakespeare]

\end{tcolorbox}

\subsection{Supplementary Prompts for JCQL}\label{appendix:prompt:supplementary}
As shown in Prompt \ref{prompts for selecting entity}, the prompt is designed to select entities that are most related to the last \texttt{thought} based on the description of entities. During the reasoning, LLM may generate ambiguous entity names. With Prompt \ref{prompts for selecting entity}, LLM will select the suitable entity.
\begin{tcolorbox}[prompt=
{\begin{prompt} \label{prompts for selecting entity}The Prompt for Entity Selection
\end{prompt}}]
$\bullet$ Instruction: \\ \\
Given entities and their descriptions, please select the entity that is most related to the thought. \\ 
\\
$\bullet$ Example: \\ \\
\textbf{Thought: }Based on the given observation, I need to find out who the leader of America is. \\
\textbf{Entity Name: }Libya \\
\textbf{Candidate Entities:} \\
Q30: country primarily located in North America \\
Q106315054: vocal track by Donna Fargo; 1974 studio recording \\
Q99281400: country of the United States of America as depicted in Star Trek \\
\textbf{Answer: }Q30
\end{tcolorbox}

As shown in Prompt \ref{prompts for selecting relations}, the prompt is designed to select relations that are most related to the last \texttt{thought} in the action \texttt{search}.

\begin{tcolorbox}[prompt=
{\begin{prompt} \label{prompts for selecting relations}The Prompt for Relation Selection
\end{prompt}}]
$\bullet$ Instruction: \\ \\
Please select 3 relations that most relevant to the thought and rank them. \\ 
\\
$\bullet$ Example: \\ \\
\textbf{Thought: }The task requires identifying educational institutions attended by Woodrow Wilson.  I will focus on gathering educational institutions associated with him. \\
\textbf{Entity Name: }Woodrow Wilson \\
\textbf{Relations:} sibling, award received, signatory, owned by, doctoral student, spouse, member of, educated at, child, occupation, described by source, archives at, topic's main category, member of political party, place of burial, successful candidate, father, residence, employer, named after, work location, writing language, position held, sex or gender, candidate, given name, family name, candidacy in election\\
\textbf{Answers:} educated at, employer, named after
\end{tcolorbox}

As shown in Prompt \ref{prompts for generating triples}, the prompt is designed to generate new triples that are most related to the sub-question in the action \texttt{generate}. \textcolor[rgb]{0.98,0.4,0.5}{The pink font} indicates the supervision in the training phase of the KBQA task.

\begin{tcolorbox}[prompt=
{\begin{prompt} \label{prompts for generating triples}The Prompt for Triple Generation
\end{prompt}}]
$\bullet$ Instruction: \\ \\
Given the existing triples, please think step by step and generate new triples related to the current question. \\
\\
$\bullet$ Example: \\ \\
\textbf{Question:} what are the types of government in the United States? \\
\textcolor[rgb]{0.98,0.4,0.5}{\textbf{Hint:} federal republic | constitutional republic | presidential system} \\
\textbf{Known Triples: }""" \\
United States of America, instance of, country \\
United States of America, instance of, constitutional republic \\
United States of America, has cabinet, United States Cabinet \\
United States of America, basic form of government, republic \\
""" \\
\textbf{Generated Triples: }""" \\
1. United States of America, instance of, federal republic \\
2. United States of America, instance of, Democratic Republic \\
3. United States of America, basic form of government, presidential system \\
""" 
\end{tcolorbox}

\renewcommand{\arraystretch}{2.0}
\begin{table*}[t!]
\centering
\begin{tabular}{>{\hspace{0pt}}m{0.1\linewidth}|>{\hspace{0pt}}m{0.85\linewidth}} 
\hline
\rowcolor[rgb]{0.9490,0.9490,0.9490} 
\textbf{input} & $\langle$ Justin\ Bieber,\ father,\ ? $\rangle$ \\
\hline
\textbf{sequence} & predict tail: Justin Bieber $|$ father \par{} entity description: Justin Bieber [Canadian singer (born 1994)] \par{} related relationship: father \par{} context: $\langle$ Justin Bieber $|$ mother $|$ Pattie Mallette $\rangle$ $\langle SEP\rangle$ $\langle$ Justin Bieber $|$ sibling $|$ Jazmyn Bieber$\rangle$ $\langle SEP \rangle$ ... \\
\hline
\rowcolor[rgb]{0.9490,0.9490,0.9490} 
\textbf{output} & Jeremy Bieber \\
\hline
\end{tabular}
\caption{A typical case of input sequences and corresponding output for the SLM.}
\label{tab:cmp_kbc}
\end{table*}
\renewcommand{\arraystretch}{1}

We first link the relations within the generated triples to the canonical relations in the background KB using the BM25 algorithm. As shown in Prompt \ref{prompts for modifying triples}, the prompt is then designed to modify the triples generated by LLM. The LLM often frequently generates unreliable triples. For instance, this prompt may allow the LLM invert head-tail entity pairs, such as modifying $\langle$Pak Pong-ju, head of government, North Korea$\rangle$ into $\langle$North Korea, head of government, Pak Pong-ju$\rangle$. Specifically, as a pre-processing step, for each relation in the background KB, the description and schema are obtained by retrieving the original description from Wikidata and extracting multiple corresponding triples from the background KB. Subsequently, the original description and these triples are input into the LLM to generate refined description and to summarize the schema from the triples.

\begin{tcolorbox}[prompt=
{\begin{prompt} \label{prompts for modifying triples}The Prompt for Triple Modification
\end{prompt}}]
$\bullet$ Instruction: \\ \\
Please modify the known triples based on the descriptions and schemas of the relations. You should generate triples without redundant words. \\
\\
$\bullet$ Example: \\ \\
\textbf{Question: }who is ruling north korea now? \\
\textbf{Known Triples: }""" \\
North Korea, office held by head of state, Kim Il-sung \\
North Korea, office held by head of state, Kim Jong-un \\
Pak Pong-ju, head of government, North Korea \\
""" \\
\textbf{Descriptions: }""" \\
1. office held by head of state: This relation links a geographical or political entity... \\
2. member of political party: This relation connects individuals to the political parties... \\
3. head of government: This relation links a specific location—such as a city, municipality... \\
4. head of state: This relation links a geographical entity, such as a country or state... \\
""" \\
\textbf{Schemas: }""" \\
1. office held by head of state: [Location], office held by head of state, [Political Office] \\
2. member of political party: [Human], member of political party, [Political Party] \\
3. head of government: [Location], head of government, [Human] \\
4. head of state: [Location], head of state, [Human] \\
""" \\
\textbf{New Triples: }""" \\
1. North Korea, head of state, Kim Il-sung \\
2. North Korea, head of state, Kim Jong-un \\
3. North Korea, head of government, Pak Pong-ju \\
"""
\end{tcolorbox}

As shown in Prompt \ref{prompts for generating paths}, the prompt is designed to select the triples most relevant to the question to construct reasoning paths.

\begin{tcolorbox}[prompt=
{\begin{prompt} \label{prompts for generating paths}The Prompt for Path Generation
\end{prompt}}]
$\bullet$ Instruction: \\ \\
Considering the known triples and relations, please select the triples and relations that are relevant to the questions and answers. \\
\\
$\bullet$ Example: \\ \\
\textbf{Question:} which one of Shakespeare's parents worked as a business person? \\
\textbf{Answer:} John Shakespeare \\
\textbf{Topic entity:} William Shakespeare \\
\textbf{Known triples:} """ \\
William Shakespeare, field of work, acting \\
William Shakespeare, field of work, poetry \\
William Shakespeare, field of work, theatre \\
William Shakespeare, father, John Shakespeare \\
William Shakespeare, mother, Mary Shakespeare \\
Mary Shakespeare, spouse, John Shakespeare \\
Mary Shakespeare, occupation, writer \\
John Shakespeare, occupation, merchant \\
John Shakespeare, occupation, politician \\
""" \\
\textbf{Related relations:} father, mother, occupation \\
\textbf{Related triples:} """ \\
William Shakespeare, father, John Shakespeare \\
William Shakespeare, mother, Mary Shakespeare \\
Mary Shakespeare, occupation, writer \\
John Shakespeare, occupation, merchant \\
John Shakespeare, occupation, politician \\
""" 
\end{tcolorbox}

\section{Case Study} \label{appendix:case study}
% \paragraph{\textbf{Comparison between GoG and JCQL under 50\% KB.}}
\paragraph{\textbf{Comparison between \texttt{generate} and \texttt{complete} under 50\% KB.}}
JCQL treats LLM as both agent and KB, generating new triples from the internal knowledge of the LLM. However, when triples are missing, action \texttt{generate} over-relies on the LLM's internal knowledge, increasing the risk of hallucination. To alleviate this issue, our approach incorporates the SLM, ensuring that predicted relations and entities are assigned unique IDs. As shown in Table \ref{tab:cmp_gog}, action \texttt{generate} incorrectly generates relationships and entities that do not exist in the KB. In contrast, JCQL generates valid entities through its action \texttt{complete}.
\paragraph{\textbf{A typical case of input sequences and corresponding output for the SLM.}} table \ref{tab:cmp_kbc} shows a typical case of input sequences and corresponding output for the SLM. Given the query $\langle Justin\ Bieber, f ather, ? \rangle$, JCQL obtains triples with the one-hop neighborhoods of the head entity $Justin\ Bieber$ uniformly, at random and without replacement. Finally, we concatenate them with $\langle SEP \rangle$.

\renewcommand{\arraystretch}{1.5}
\begin{table*}[t!]
\centering
\begin{tabular}{>{\hspace{0pt}}m{0.1\linewidth}|>{\hspace{0pt}}m{0.83\linewidth}} 
\hline
\rowcolor[rgb]{0.8588,0.8588,0.8588} Question   & what type of government did soviet union have? \\ 
\hline
\rowcolor[rgb]{0.9490,0.9490,0.9490} JCQL w/o \texttt{complete}   & Thought 1:~I need to find out what type of government the Soviet Union had.\par{}Action 1:~Search[Soviet Union]\par{}Observation 1:~soviet union, history of topic, history of the soviet union\par{}soviet union, legislative body, supreme soviet of the soviet union\par{}soviet union, official religion, state atheism\par\null\par{}Thought 2:~The Soviet Union had a legislative body called the Supreme Soviet, which implies a type of government. I need to generate more information about the government structure of the Soviet Union based on this observation.\par{}Action 2:~\textcolor[rgb]{0.7529,0,0}{Generate[type of government of the Soviet Union]}\par{}Observation 2: soviet union, ideological basis, Marxism-Leninism\par{}soviet union, type of government, one-party socialist republic\par\null\par{}Thought 3:~The Soviet Union is characterized as a one-party socialist republic based on its ideological basis of Marxism-Leninism.\par{}Action 3:~Finish[one-party socialist republic]  \\
\hline
\rowcolor[rgb]{0.9490,0.9490,0.9490} JCQL w/o \texttt{generate}                     & Thought 1:~The question requires identifying the type of government in the Soviet Union. I should gather information regarding the political structure and governance of the Soviet Union.\par{}Action 1:~Search[Soviet Union]\par{}Observation 1:~Soviet Union, foundational text, Declaration of the Creation of the Union of Soviet Socialist Republics\par{}Soviet Union, foundational text, Treaty on the Creation of the Union of Soviet Socialist Republics\par{}Soviet Union, legislative body, Supreme Soviet of the Soviet Union\par{}member of the Supreme Soviet of the Russian Soviet Federative Socialist Republic, country, Soviet Union\par\null\par{}Thought 2:~The current information provides some context about the Soviet Union but lacks specific details about its type of government. I need to find the governance structure of the Soviet Union.\par{}Action 2:~\textcolor[rgb]{0.4549,0.7137,0.0078}{Complete[Soviet Union | instance of]}\par{}Observation 2: Soviet Union, instance of, country\par{}Soviet Union, instance of, \textcolor[rgb]{0,0.4392,0.7529}{socialist state}\par\null\par{}Thought 3:~The results indicate that the Soviet Union was a socialist state. This clarifies the type of government. Therefore, I can summarize the answers.\par{}Action 3:~Finish[\textcolor[rgb]{0,0.4392,0.7529}{socialist state}]. \\
\hline
\end{tabular}
\caption{The comparison between \texttt{generate} and \texttt{complete} under the setting i.e., 50\% KB.}
\label{tab:cmp_gog}
\end{table*}
\renewcommand{\arraystretch}{1}

% \section{Supplementary Related Work} 

\section{Supplementary Setting Details} \label{appendix:implementation_details}
In all experiments, the depth of exploration is set to $10$ to avoid endless exploration. The maximum token length for generation is set to $512$. All experiments are conducted on a server with an Intel(R) Xeon(R) Silver $4310$ CPU ($2.10$GHz), $256$ GB RAM memory, and an NVIDIA RTX A6000 GPU.

\section{Broader Impact} \label{appendix:limitations}
By performing KBQA and KBC jointly, JCQL enhances the complex reasoning abilities of the KB-augmented LLM and improves the performance of the SLM when faced with incomplete KBs. Furthermore, other broader impacts of JCQL are listed as follows. (1) The proposed framework JCQL is flexible. For KBQA, we can choose any KBC model to complete missing triples for any KB. 
(2) JCQL can generate reasoning paths based on input questions and update the KB in real time using incremental learning techniques, demonstrating its real-time performance and the ability of lifelong learning.
(3) From a long-term perspective, the LLM's reasoning paths generated from continuous increasing of question-answer pairs strengthen the KBC model, which in turn improves the reasoning capability of LLM.

\end{document}